\documentclass[lettersize,journal]{IEEEtran}
\usepackage{amsmath,amsfonts}
\usepackage{algorithmic}
\usepackage{array}
\usepackage{romannum}
\usepackage[caption=false,font=normalsize,labelfont=sf,textfont=sf]{subfig}
\usepackage{pifont}
\usepackage{hyperref}
\usepackage{array}         
\usepackage[dvipsnames, svgnames]{xcolor} 
\definecolor{mylightblue}{HTML}{367DBD}
\definecolor{mylightred}{HTML}{FF0000}
\hypersetup{
    colorlinks = true,   
    linkcolor = mylightred,    
    citecolor = CornflowerBlue,    
    urlcolor = blue      
}

\usepackage{textcomp}
\usepackage{cleveref} 
\usepackage{cite}
\usepackage{amsmath}
\usepackage{colortbl}
\usepackage{stfloats}
\usepackage{booktabs}
\usepackage{url}
\usepackage{verbatim}
\usepackage{multirow}
\usepackage{graphicx}
\def\BibTeX{{\rm B\kern-.05em{\sc i\kern-.025em b}\kern-.08em
    T\kern-.1667em\lower.7ex\hbox{E}\kern-.125emX}}
\usepackage{balance}

\renewcommand{\thesubsubsection}{\thesubsection-\arabic{subsubsection}}
\begin{document}
\title{PTMs-TSCIL: Pre-Trained Models Based Class-Incremental Learning for Time Series}

\author{Yuanlong Wu, 
Mingxing Nie\textsuperscript{*}, ~\IEEEmembership{Member,~IEEE,} 
Tao Zhu, ~\IEEEmembership{Senior Member,~IEEE,} 
Liming Chen, ~\IEEEmembership{Senior Member,~IEEE,} Huansheng Ning, ~\IEEEmembership{Senior Member,~IEEE,} 
Yaping Wan, ~\IEEEmembership{Member,~IEEE,}

\thanks{Yuanlong Wu, Mingxing Nie, Tao Zhu and Yaping Wan are with the School of Computer Science, University of South China, 421001 China
(e-mail: ylwu@stu.usc.edu.cn, niemx@usc.edu.cn, tzhu@usc.edu.cn, ypwan@aliyun.com). (Corresponding author: Mingxing Nie.)

Liming Chen was with School of Computer Science and Technology, Dalian University of Technology, 116024 China.
Huansheng Ning was Department of Computer \& Communication Engineering, University of Science and Technology Beijing, 100083 China. (email: limingchen0922@dlut.edu.cn, ninghuansheng@ustb.edu.cn).}}

\markboth{Journal of \LaTeX\ Class Files,~Vol.~18, No.~9, September~2020}%
{How to Use the IEEEtran \LaTeX \ Templates}
\maketitle

\begin{abstract}
Class-incremental learning (CIL) for time series data faces critical challenges in balancing stability against catastrophic forgetting and plasticity for new knowledge acquisition, particularly under real-world constraints where historical data access is restricted. While pre-trained models (PTMs) have shown promise in CIL for vision and NLP domains, their potential in time series class-incremental learning (TSCIL) remains underexplored due to the scarcity of large-scale time series pre-trained models. 
Prompted by the recent emergence of large-scale pre-trained models (PTMs) for time series data, we present the first exploration of PTM-based Time Series Class-Incremental Learning (TSCIL). Our approach leverages frozen PTM backbones coupled with incrementally tuning the shared adapter, preserving generalization capabilities while mitigating feature drift through knowledge distillation. Furthermore, we introduce a Feature Drift Compensation Network (DCN), designed with a novel two-stage training strategy to precisely model feature space transformations across incremental tasks. This allows for accurate projection of old class prototypes into the new feature space. By employing DCN-corrected prototypes, we effectively enhance the unified classifier retraining, mitigating model feature drift and alleviating catastrophic forgetting.
Extensive experiments on five real-world datasets demonstrate state-of-the-art performance, with our method yielding final accuracy gains of 1.4\%–6.1\% across all datasets compared to existing PTM-based approaches.
Our work establishes a new paradigm for TSCIL, providing insights into stability-plasticity optimization for continual learning systems.

\end{abstract}

\section{Introduction}
\IEEEPARstart{T}{ime} series (TS) data are widely used in various fields, such as acoustics, healthcare, and manufacturing\cite{ruiz2021great}. Traditional deep learning methods \cite{mohammadi2024deep} assume that training data for time series classification tasks is static, and all class data follow an independent and identically distributed (i.i.d.) assumption. However, real-world data streams often exhibit dynamic changes, with continuously shifting underlying distributions. For instance, in human activity recognition tasks, models need to adapt to newly introduced activity classes while retaining knowledge of previously learned ones\cite{mahmoud2022multi,chauhan2020contauth}. This problem is known as class-incremental learning (CIL)\cite{rebuffi2017icarl,van2019three}, where a model must continually learn new classes while preserving previously acquired knowledge. The extension of class-incremental learning to time series data is referred to as time series class-incremental learning (TSCIL).

\begin{figure}[!t]
\centering
\includegraphics[width=3.45in]{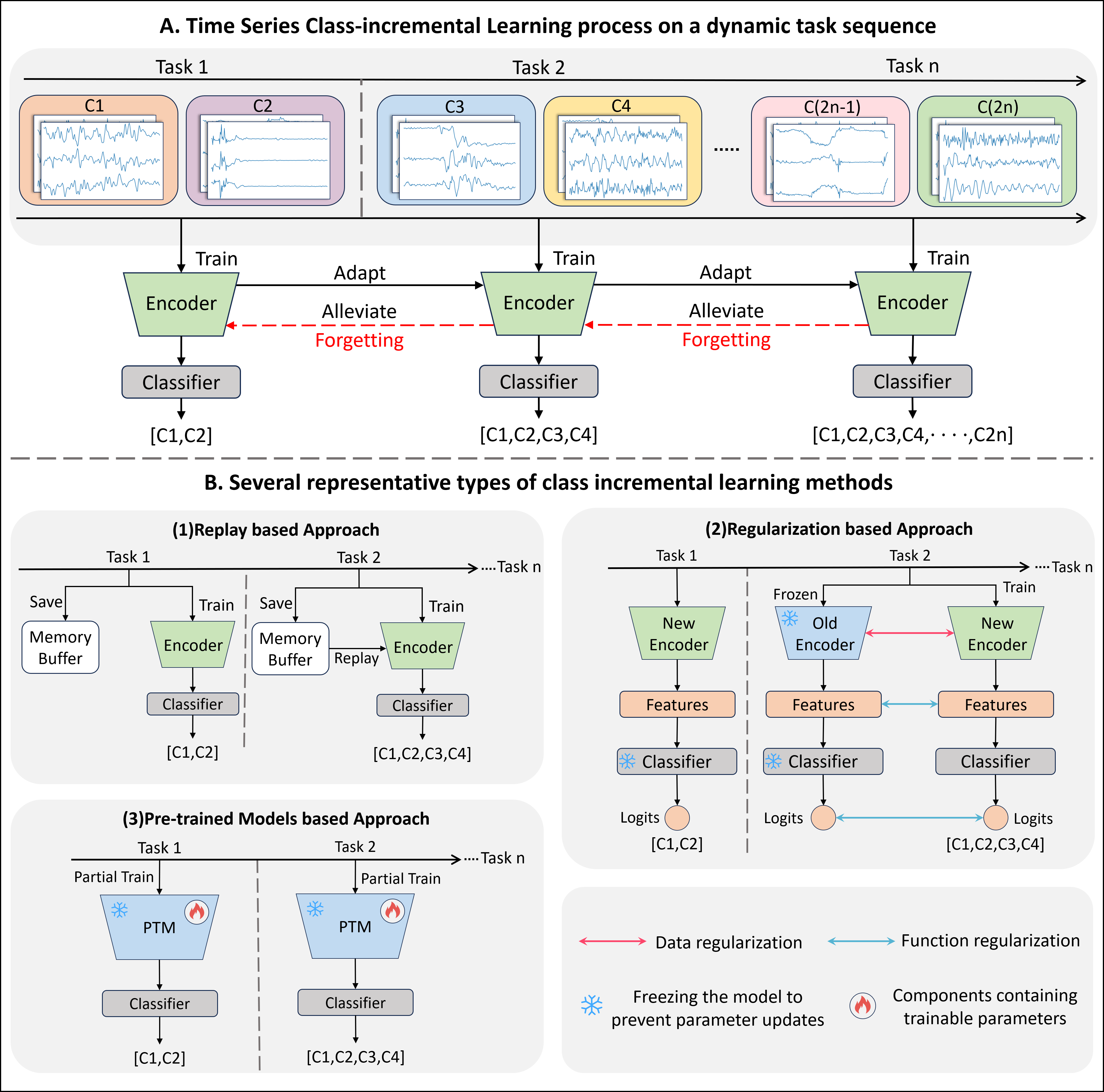}
\caption{TSCIL process schematic diagram and several representative method structure diagrams.}
\label{fig1}
\vspace{-0.3cm}
\end{figure}

A core challenge in CIL is catastrophic forgetting\cite{catastrophic}, primarily caused by the feature extractor adapting to new task distributions. This adaptation alters the feature space, shifting representations of old classes and exacerbating forgetting. However, mitigating catastrophic forgetting often compromises model plasticity. Balancing stability (retaining old knowledge) and plasticity (learning new tasks) is critical for CIL performance, making the stability-plasticity dilemma\cite{studies} a central challenge in CIL/TSCIL (see Fig.~\ref{fig1}(A)).

Researchers have proposed various methods to address these challenges. Replay-based approaches\cite{chaudhry,rolnick} mitigate forgetting by storing and replaying old class samples during training (Fig. \ref{fig1}(B-1)). However, they face practical issues like data privacy concerns and storage overhead. In contrast, Non-Exemplar Class Incremental Learning (NECIL)\cite{lwf,re12,re13} assumes no access to old class samples, aligning better with real-world scenarios. Yet, NECIL suffers from severe feature drift and catastrophic forgetting due to the lack of prior knowledge.
A classical NECIL approach is the regularization based method (Fig. \ref{fig1}(B-2)), which introduces explicit regularization terms to balance old and new tasks, often requiring frozen old model as references. Depending on regularization objectives, These methods are categorized as weight or function regularization. Knowledge distillation (KD) is widely adopted in function regularization, mitigating catastrophic forgetting by transferring knowledge from old models to new ones, yet faces challenges in insufficient plasticity \cite{re25,re26}.

Recent studies have explored pre-trained models (PTMs) for CIL. Empirical evidence suggests that PTM-based approaches are nearing the upper bound of incremental learning's potential\cite{re28}. PTMs inherently generalize well across diverse tasks, circumventing limitations of traditional models (e.g., data privacy risks, high training complexity, and poor generalization). Consequently, PTM-based CIL is gaining prominence. Typical methods\cite{re15,re16,re17slca}  freeze PTM parameters while incrementally fine-tuning lightweight modules—such as prompts \cite{re18}, adapter\cite{re19}, or SSF\cite{re20}—effectively balancing the stability-plasticity dilemma\cite{studies}. This strategy is termed parameter-efficient tuning (PET).

\begin{figure}[!t]
\centering
\includegraphics[width=3.5in]{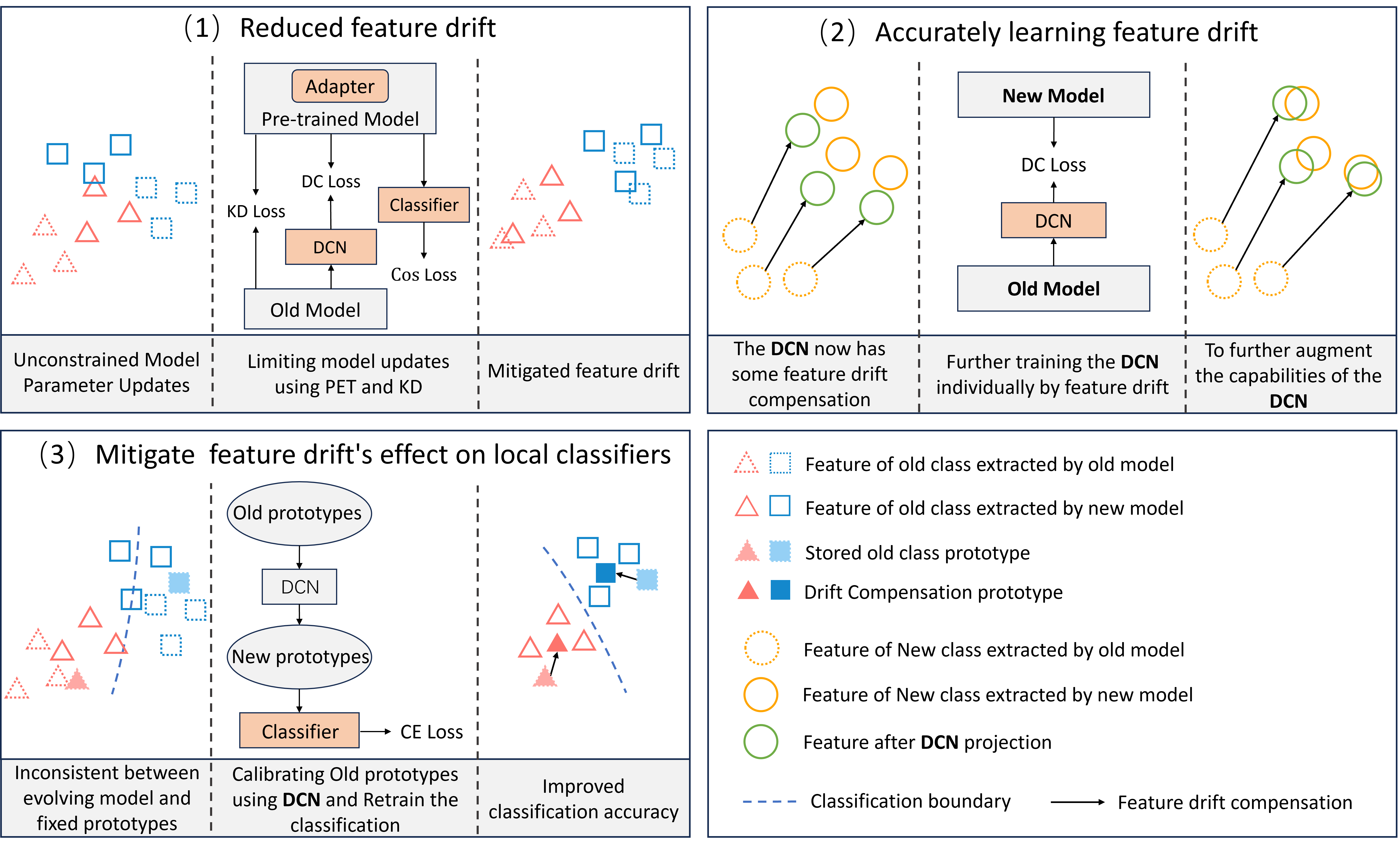}
\caption{TSCIL process schematic diagram and several representative method structure diagrams.}
\label{fig2}
\vspace{-0.3cm}
\end{figure}

Despite significant progress in PTM-based CIL for image and natural language processing (NLP), research in the time series domain remains underexplored. The primary obstacle is the lack of large-scale, well-integrated public time series datasets, making it difficult to train high-quality PTMs for time series data. However, with the recent emergence of large-scale time series pre-trained models, such as Timer\cite{re22timer}, MOIRAI\cite{re23}, and Moment\cite{re24moment}, we recognize the potential of leveraging PTMs in time series data to significantly enhance TSCIL performance. By keeping the pre-trained model frozen while progressively training additional modules (e.g., adapters) for parameter-efficient tuning (PET), we address three key challenges and propose corresponding solutions:

\textbf{Mitigating feature drift:} 
Unconstrained parameter updates in class-incremental learning (CIL) induce excessive over-updating of model parameters, exacerbating feature drift and compromising stability\cite{re27,re28} (Fig. \ref{fig2}(1-left)). To mitigate this, we implement parameter-efficient tuning (PET) by embedding trainable adapter\cite{re19} into frozen pre-trained models (PTMs), reducing trainable parameters while implicitly restricting updates. However, unregulated adapter turning may still lead to over-adaptation. We thus integrate knowledge distillation (KD)\cite{re66kd}  to regularize adapter updates, suppressing feature drift and preserving stability. Experiments validate that the adapter-KD synergy balances stability-plasticity while enhancing task accuracy (see Section \ref{sec:kd}).
Additionally, a feature drift compensation network (DCN) is preliminarily trained to learn feature mapping relationships between old and new models. The unified framework (Fig. \ref{fig2}(1-middle)) combining PET, KD, and DCN effectively minimizes feature drift (Fig. \ref{fig2}(1-right)).

\textbf{Accurately Estimating Feature Drift:} Existing studies \cite{re29} estimate feature drift across different CIL stages but neglect intra-task variations and feature-level adjustments. Research\cite{re30fcs,re53} suggests that a simple, learnable feature drift calibration network is a more effective drift compensation strategy. During parameter-efficient tuning (PET), we train a Drift Compensation Network (DCN) to capture differences between the frozen model from task t-1 and the updated model at task t (Fig. \ref{fig2}(1-middle)). After initial training, we freeze both t-1 and t models,  further refine the DCN independently (Fig. \ref{fig2}(2-middle)), to enhance its effectiveness in feature drift compensation (Fig. \ref{fig2}(2-right). This approach is analyzed in depth in Section \ref{sec:subDCN}.

\textbf{Addressing Feature Drift Impact on Local Classifiers:} Local classifiers isolate different tasks during continual learning, reducing catastrophic forgetting and enhancing task adaptability\cite{lwf}. We initialize a separate local classifier for each task, training only the current classifier while keeping previous ones frozen. However, direct testing with frozen old classifiers is suboptimal due to feature drift. Classifier re-training\cite{re12,re17slca,re31} partially mitigates this issue, but suffers from prototype misalignment across tasks\cite{re32} (Fig. \ref{fig2}(3-left)). Using the trained DCN, we correct old class prototypes within the new feature space, enhancing classifier re-training performance (Fig. \ref{fig2}(3-right)).

In summary, the main contributions of this article are as follows:
\begin{enumerate}
\item First exploration of PTMs-based CIL in the time series domain, proposing a novel framework integrating continual adapter tuning and feature drift learning.
\item Design a two-stage training strategy for the Drift Compensation Network (DCN) to model feature mapping between old and new models. This enables precise calibration of old class prototypes, generating reliable feature samples for unified classifier retraining, effectively mitigating catastrophic forgetting.

\item Comprehensive experiments on five time series datasets and multiple benchmarks, demonstrating our model’s superior performance across various task settings. Our results offer novel insights and effective methodologies for future TSCIL research.
\end{enumerate}

\section{Related Work}
\subsection{Class-Incremental Learning} \label{sec:CIL}
Class-incremental learning (CIL) requires models to learn new class instances while retaining knowledge of previously learned classes\cite{re28}. Traditional CIL methods can be categorized into the following approaches: replay-based\cite{re33,chaudhry,rolnick}, regularization-based\cite{lwf,re34,re12}, and parameter-isolation-based\cite{re35,re36} methods.
Replay-based methods store or generate old class samples and incorporate them into the current training process. These methods employ sample selection or generation strategies to effectively replay past information, mitigating catastrophic forgetting. 
Regularization-based methods introduce constraints or penalties during learning to restrict updates to important parameters related to old classes. This is typically achieved by adjusting the loss function to retain knowledge of previously learned tasks. 
Parameter-isolation methods dynamically adjust network architecture or isolate task-specific parameters, selectively updating subsets to prevent interference and minimize catastrophic forgetting.
Despite the progress made, replay-based methods and most regularization-based or parameter-isolation-based approaches require data storage, which raises significant concerns regarding data privacy.

\subsection{Class-Incremental Learning for Time Series}
In existing studies on time series class-incremental learning (TSCIL), most approaches follow conventional CIL methods, particularly exemplar replay (ER) based methods. For example, CLOPS\cite{re37} is an ER-based arrhythmia diagnosis approach that employs importance-based sample storage and uncertainty-based memory buffer management strategies. 
Additionally, regularization-based methods have been widely applied. The study in\cite{chauhan2020contauth} proposed an online user authentication framework combining EWC\cite{re38} and iCaRL\cite{rebuffi2017icarl} for continuous user identification based on biomedical time series signals. while\cite{re40} used recurrent neural network (RNN) to evaluate various CIL methods on simple time series datasets such as Stroke-MNIST\cite{re41} and AudioSet\cite{re42}, showing general CIL methods can partially mitigate catastrophic forgetting in time series data.
Outside of general CIL methods, Recent work has introduced innovative CIL algorithms tailored for time series data. DT2W\cite{re43} proposes a novel soft-DTW\cite{re44} based knowledge distillation (KD) strategy that compares teacher and student model feature maps to improve incremental learning. Meanwhile, MAPIC\cite{re45}, a meta self-attention prototype incremental classifier, integrates meta self-attention, prototype augmentation, and KD to tackle few-shot class-incremental learning (FSCIL) for medical time series.
studies\cite{re46TSCIL} establishes the first systematic benchmark for time series class-incremental learning (TSCIL), providing standardized evaluation protocols and comparative analysis of existing methods. Our experimental design aligns with this benchmark to ensure methodological consistency.

\subsection{Class-Incremental Learning on Pre-Trained Models}
Unlike traditional CIL methods, PTMs-based approaches leverage large-scale pre-trained models for continual learning instead of training from scratch. The study\cite{re28} shows that PTMs-based methods achieve performance close to the upper bound of continual learning potential, making them promising for real-world applications.
Most existing PTMs-based CIL methods employ pre-trained Vision Transformers (ViTs)\cite{re47} as feature extractors. 
Representation-based approaches directly leverage the PTM's generalization capacity to construct classifiers.
SLCA\cite{re17slca} explores the fine-tuning paradigm of the PTM's, setting different learning rates for backbone and classifiers, and gains excellent performance.
Adam\cite{re49} proposes to construct the classifier by merging the embeddings of a pre-trained model and an adapted downstream model. 
Model mixture-based methods design ensembles of models during learning, using model merging, ensembling, or hybrid techniques for final predictions. 
SSIAT\cite{re51} performs unconstrained incremental adaptation of shared adapters, concurrently estimating semantic drift in old prototypes and retrain the classifier using updated prototypes in each session.

\begin{table}
\setlength{\abovecaptionskip}{0cm} 
\setlength{\belowcaptionskip}{0cm}
\setlength{\tabcolsep}{3.2pt} 
\begin{center}
\caption{Overview of Several Large Time Series Pre-training Models}
\label{tab1}
\begin{tabular}{@{}cccc@{}}
\toprule
Method                          & \textbf{TIMER}         & \textbf{MOIRAI}                    & \textbf{MOMENT}              \\ \midrule
Architecture                    & Decoder       & Encoder                   & Encoder             \\
Model Size                      & 29M, 50M, 67M & 14M, 91M, 311M            & 40M, 125M, 385M     \\
\multirow{3}{*}{Supported Task} & Forecast      & \multirow{3}{*}{Forecast} & Forecast Imputation \\[-0.5em] 
                                & Imputation    &                           & Detection           \\[-0.5em] 
                                & Detection     &                           & Classification      \\
Pre-training Scale              & 28B           & 27.65B                    & 1.13B               \\ \bottomrule
\end{tabular}
\end{center}
\vspace{-0.5cm}
\end{table}

\subsection{Large Time Series Pre-Trained Models}
The lack of large-scale, integrated public time series datasets has hindered the development of high-performance large time-series models. Recently, researchers have integrated datasets from multiple domains to develop large-scale time-series pre-trained models, paving the way for new research opportunities. Below, we introduce several recent open-source time series foundation models, summarized in Tab. \ref{tab1}.

Timer\cite{re22timer}: A GPT-based generative pre-trained model trained on diverse, large-scale time series datasets using autoregressive pre-training mechanisms. The dataset spans finance, meteorology, healthcare, and more, ensuring broad applicability. Timer demonstrates superior performance in forecasting, imputation, and anomaly detection tasks, surpassing traditional methods with strong generalization and multi-task adaptability.

MOIRAI\cite{re23}: Introduces U-TFT, a unified pre-training framework for training general time series forecasting Transformers. U-TFT leverages multi-task learning across reconstruction, forecasting, and classification objectives to learn cross-domain general time-series representations. Pre-trained on energy, transportation, and healthcare datasets, MOIRAI significantly improves forecasting accuracy across multiple benchmarks.

Moment\cite{re24moment}: A general-purpose time-series foundation model trained on the Time Series Pile, a diverse public time-series dataset. Moment employs patch-based encoding with random masked embedding techniques to enhance feature learning. Evaluations on 91 time-series datasets demonstrate Moment’s exceptional representation learning capabilities, particularly in classification tasks, outperforming specialized time-series models without fine-tuning.


\section{Methods}
\subsection{Problem Setting}
In the definition of class incremental learning, the entire data stream is partitioned into a sequence of $T$ tasks, denoted as
$\mathcal{D}=\{ \mathcal{D}^{1}, \mathcal{D}^{2}, \dots, \mathcal{D}^{T} \}$, where
$\mathcal{D}^{t}=\{ X_{t}, Y_{t} \}$
is the t-th incremental task with the training set
$X_{t} = \{ x_{t,j} \mid j = 1, \dots, n_{t} \}$
and the label set
$Y_{t} = \{ y_{t,j} \mid y_{t,j} \in \mathcal{C}_{t}, \ j = 1, \dots, n_{t} \}$
where $n_{t}$ is the number of samples in stage $t$ , $x_{t,j}$ denotes the j-th time series sample with shape $\mathbb{R}^{C \times L}$, where $C$ denotes the number of channels/variables and $L$ denotes the length of the sequence. ${C}_{t}$ is the label set of the stage ${t}$. 
When $t \neq t^{'}$ we all have $\mathcal{C}_{t} \cap \mathcal{C}_{t'} = \emptyset$, i.e., in the t-th training task, we can only access the data in $\mathcal{D}^{t}$ for model updating, and the aliases in each task do not overlap.

The goal of CIL is to gradually construct a unified model $\mathcal{M} = f_{\theta_{\text{cls}}} (\mathcal{F}(\cdot))$ for all visible classes
which is capable of acquiring knowledge from new classes while retaining knowledge from previous classes. Where $\mathcal{F}(\cdot)$ denotes the backbone network that extracts features from the input samples, and $f_{\theta_{\text{cls}}}$ is the categorization layer with the parameter ${\theta_{\text{cls}}}$. After each incremental task (e.g., task $t$), the ability of the model  $\mathcal{M}$ is evaluated on all visible categories $\mathcal{Y}_{t} = Y_{1}\cup\cdot\cdot\cdot\cup{Y}_{t}$. Formally, the goal is to fit a model $\mathcal{M}(x) : X \to \mathcal{Y}_{t}$, minimizing the empirical risk across all test data sets:

\begin{equation}
\label{deqn_e1}
\min \sum_{(x_j, y_j) \in \mathcal{D}_{\text{test}}^{1} \cup \dots \cup \mathcal{D}_{\text{test}}^{t}} \ell(\mathcal{M}(x_j), y_j),
\end{equation}

where $\mathcal{\ell}(\cdot, \cdot)$ measures the difference between the predicted and true label. $\mathcal{D}_{\text{test}}^{t}$ denotes the test set for task $t$. A TSCIL model that satisfies Eq. (\ref{deqn_e1}) is discriminative for all categories and can strike a balance between learning new categories and memorizing old ones.

\begin{figure*}[!t]
\centering
\includegraphics[width=5.5in]{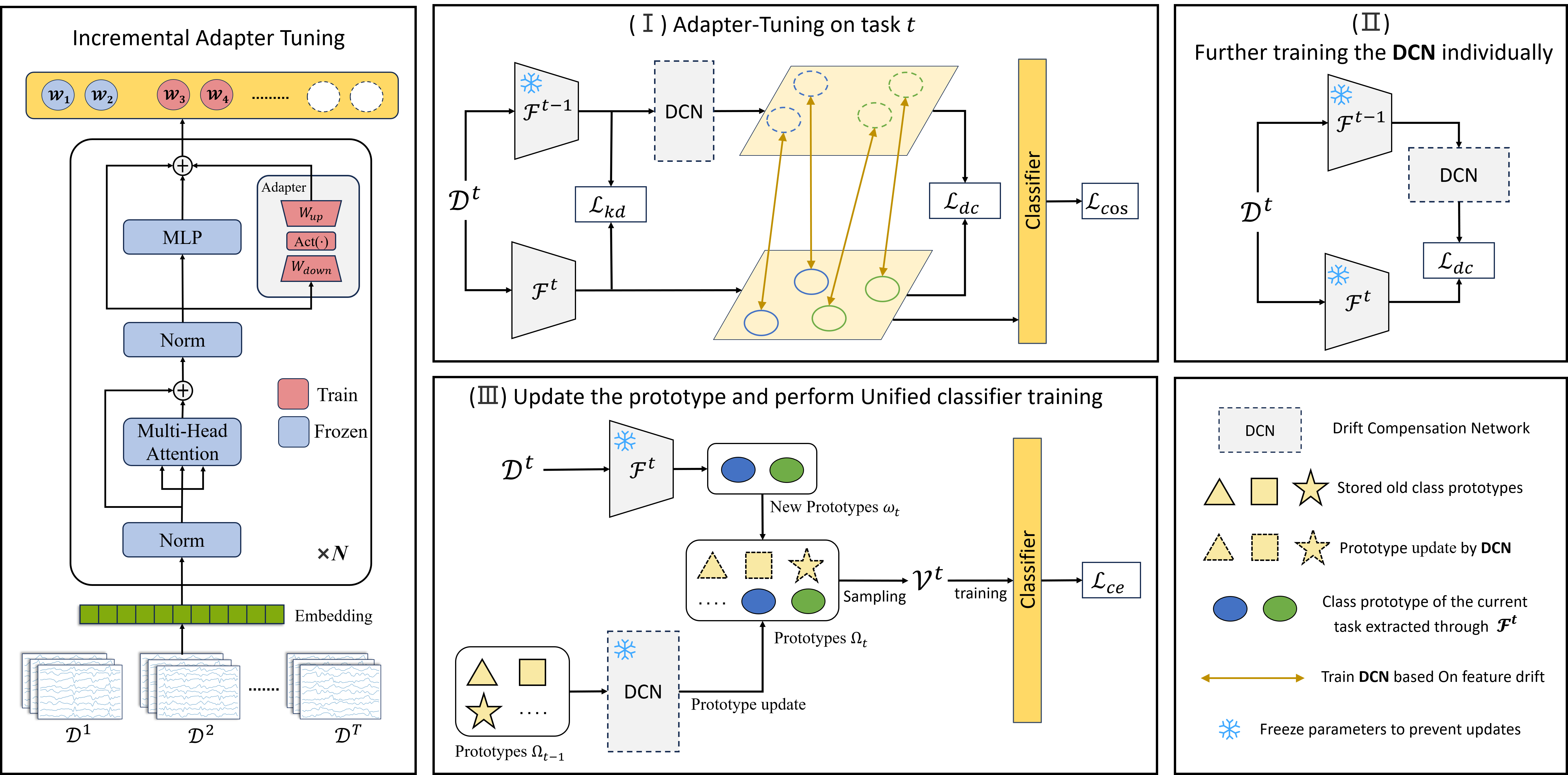}
\caption{The framework of our proposed method. \textbf{Left:}The structure diagram of the Moment, Adapter, and local classification head. 
\label{fig3}
\textbf{Right:}The training process comprises these steps:
(I) In task \(t\), the Adapter and local classifier are trained, and the Drift Compensation Network (DCN) is preliminarily trained.  
(II) The DCN is further trained separately by leveraging feature drift between the old and new models.  
(III) Drift compensation is applied to the old class prototypes, and new class prototypes are extracted and stored. Subsequently, samples are generated using these prototypes, and the classification head is retrained.  
}
\vspace{-0.35cm}
\end{figure*}

\subsection{Continual Adapter Tuning}
Most methods based on Pre-trained Models (PTMs) typically keep the PTMs frozen while progressively training shared additional modules, such as prompts, adapters, and SSF\cite{re65scaling} (Scalable \& Stable Fine-tuning), to achieve better performance. However, SSIAT\cite{re51} demonstrates that adjusting the same prompt parameters during each learning process can lead to catastrophic forgetting. Compared to prompts and SSF, adapter are more suitable for continual learning. Gradually fine-tuning shared adapters can effectively mitigate the conflict between learning new categories and retaining knowledge of old categories.

The adapter is a bottleneck structure\cite{chauhan2020contauth} integrated into transformer-based PTMs. By freezing PTMs weights and training only the adapter, parameter updates are minimized, reducing interference with existing knowledge and improving efficiency. The adapter consists of a down-sampling layer, a non-linear activation function (e.g., ReLU), and an up-sampling layer. Given an input $x_i$, The adapter output is expressed as:

\begin{equation}
\label{deqn_e2}
out = x_i + \text{RELU}(s \cdot (x_i * W_{\text{down}})) * W_{\text{up}}
\end{equation}

where $s$ denotes the scale factor, and $*$ denotes the matrix multiplication.

The adapter tuning method in SSIAT\cite{re51} avoids knowledge distillation and instead constrains model updates by applying a small scale factor to the Adapter. In contrast, we propose that combining an appropriate Adapter scale factor with knowledge distillation \(\mathcal{L}_{kd}\) (Eq. (\ref{deqn_e3})) during shared Adapter training better balances plasticity and forgetting. (See Section \ref{sec:kd}) Furthermore, we introduce the DCN training loss \(\mathcal{L}_{dc}\) to reduce feature drift between old and new models and facilitate DCN learning of such drift. Each task's training consists of three stages. For instance, in the first stage of task \(t\), the shared Adapter, DCN, and local classifiers are jointly trained using \(\mathcal{L}_{kd}^t\), \(\mathcal{L}_{dc}^t\), and \(\mathcal{L}_{cos}^t\), as shown in Fig. \ref{fig3}(I). Both \(\mathcal{L}_{kd}^t\) and \(\mathcal{L}_{dc}^t\) are constructed based on Euclidean distance, where \(\mathcal{L}_{dc}^t\) is shown in Eq. (\ref{deqn_e5}) and \(\mathcal{L}_{kd}^t\) is represented as follows:

\begin{equation}
\label{deqn_e3}
\mathcal{L}_{kd}^{t} = \frac{1}{N} \sum_{i=1}^{N} \left\| \mathcal{F}^{t-1}(x_i) - \mathcal{F}^{t}(x_i) \right\|_2^2
\end{equation}

As the cosine classifier has shown great success in CIL, we follow\cite{re52} to use the cosine classifier with a margin. The training loss can be formulated as follows:

\begin{equation}
\label{deqn_e4}
\mathcal{L}_{cos}^t=-\frac{1}{N^t}\sum_{j=1}^{N^t}log\frac{e^{s(cos\theta_j^i-m)}}{e^{s\left(cos\theta_j^i-m\right)}+\sum_{c=1}^{Y(t)-i}e^{s\left(cos\theta_j^c\right)}}
\end{equation}

\( \cos \theta_{j}^{i}=\frac{w_i * f_j}{\| w_i \| * \| f_j \|} \), where $w_i$ and $f_j$ are the weight vector of class $i$ (classification head) and the feature vector of sample $j$, respectively.
${N^t}$ denotes the number of training samples for the current task, and $s$ and $m$ denote the scale factor and margin factor, respectively.
Since we do not retain training samples from old classes, gradients computed on current samples affect both new and previously learned classifiers, leading to significant forgetting. To mitigate this, we adopt a local training loss strategy inspired by prior works\cite{re50,re17slca,re51}. Specifically, we initialize a new classification head at the start of each task, compute the loss only between current logits and labels, and block gradient updates to previous classifiers. This approach alleviates forgetting by isolating the optimization of current task parameters. The classification heads are illustrated in the top left of Fig. \ref{fig3}.

\subsection{Drift Compensation Network}
As new tasks are continually learned, the trainable parameters of the model inevitably change, leading to feature distribution drift for old classes. Directly using stored old class prototypes for unified classification head training may result in misclassification of old classes during testing. To address this, traditional methods often employ Semantic Drift Compensation (SDC), which approximates prototype drift by computing feature drift between new and old models on new task data. While SDC mitigates forgetting without accessing historical samples, recent studies\cite{re30fcs} suggest that Learnable Drift Compensation network (DCN) outperforms SDC. SDC assumes drift follows a fixed local transformation (e.g., translation), but this may introduce estimation errors if drift follows other transformations (e.g., scaling).

To accurately learn feature drift, we use the current task dataset \(\mathcal{D}_t\) to train a DCN by minimizing the Euclidean distance between the projected features of \(\mathcal{F}^t(\cdot)\) and \(\mathcal{F}^{t-1}(\cdot)\). Unlike training DCN solely during adapter tuning (Fig. \ref{fig3}(I)) or after adapter tuning (Fig. \ref{fig3}(II)), we propose a two-stage DCN training strategy: preliminary DCN training followed by dedicated refinement, which significantly improves drift compensation. (Detailed results are discussed in Section \ref{sec:subDCN}). Both stages share the same loss function:

\begin{equation}
\label{deqn_e5}
\mathcal{L}_{dc}^t = \frac{1}{N} \sum_{i=1}^{N} \| d^t(\mathcal{F}^{t-1}(x_i)) - \mathcal{F}^t(x_i) \|^2
\end{equation}

where \(d^t(\cdot)\) denotes the DCN in task $t$, \(\mathcal{F}^{t-1}(\cdot)\) remains frozen. In the first stage, we jointly train the adapter, classification head, and DCN. In the second stage, we freeze both \(\mathcal{F}^{t-1}(\cdot)\) and \(\mathcal{F}^t(\cdot)\), focusing solely on refining the DCN. After both stages, the stored prototypes are updated using the DCN and merged with new class prototypes \(p^t\) (Eq. (\ref{deqn_e6})) to form the updated prototype set \(P^t = \{d^t(P^{t-1})\} \cup p^t\).

\begin{equation}
\label{deqn_e6}
\begin{split}
p^t = \{p_1^t, p_2^t, \dots, p_{|\mathcal{C}_t|}^t\}, \\
p_i^t = \frac{1}{|\mathcal{D}_{y_i}^t|} \sum_{(x_j, y_j) \in \mathcal{D}_{y_i}^t} \mathcal{F}^t(x_j)
\end{split}
\end{equation}

where \(\mathcal{D}_{y_i}^t = \{(x_j, y_j) \mid y_j = y_i, (x_j, y_j) \in \mathcal{D}^t\}\) and \(|\mathcal{C}_t|\) denotes the number of classes in \(\mathcal{D}^t\).

For the DCN architecture, studies\cite{re53} show that a bias-free linear layer achieves optimal performance. Thus, we employ a bias-free linear layer as the core structure of the DCN. To enhance stability in learning feature drift, we initialize the DCN weights as an identity matrix at the start of each task.

After the first task, this process is repeated for each new task. Throughout training, the DCN is trained \(T-1\) times, continuously updating prototypes to adapt to the evolving feature space of the model.

\subsection{Unified classifier training}
To mitigate classifier forgetting, we employ a local training loss that prevents gradient updates to previous classifiers, focusing solely on training the current task's classifier. Although freezing old classifiers reduces direct interference, the trainable parameters of the model still change as new tasks are learned, leading to performance degradation and catastrophic forgetting if prior classifiers remain unadjusted. Previous works \cite{re17slca,re12,re51} address this by modeling each class as a Gaussian distribution and sampling features to retrain a unified classifier. We adopt a similar approach, integrating a DCN to correct old class prototypes for unified classifier training (Fig. \ref{fig3}(III)).

Specifically, after training each task, we compute the class prototypes and covariance matrices for the involved classes. Given that the trained backbone network provides uniformly distributed representations, each class exhibits a unimodal distribution, modeled as a Gaussian with prototype \(\mu_c\) and covariance matrix \(\Sigma_c\). Feature samples \(\mathcal{V}_c\) for class \(c\) are sampled from this distribution, forming the feature sample set \(\mathcal{V}^t = \bigcup_{c=1}^{C} \mathcal{V}_c\) for all visible classes \(C\) at task \(t\). Each \(\mathcal{V}_c\) is generated as:

\begin{equation}
\label{deqn_e7}
\mathcal{V}_c = \{v_{c,1}, \dots, v_{c,S_n}\}, \quad v_{c,i} = \mu_c + \Sigma_c^{1/2} z_i,
\end{equation}

where\(\ i = 1, 2, \dots, S_n\), \(S_n\) denotes the number of sampled instances per class, and \(\mathbf{z}_i \sim \mathcal{N}(0, I)\) represents a noise vector sampled from a standard normal distribution, with a mean of 0 and an identity covariance matrix \(I\).

Using the sample set \(\mathcal{V}^t\), we train the classification layer \(f(\cdot)\) with parameters \(\theta_{cls}\) via cross-entropy loss:

\begin{equation}
\label{deqn_e8}
\mathcal{L}_{ce}(\theta_{cls}, \mathcal{V}^t) = -\sum_{i=1}^{S_n \cdot C} \log \frac{e^{\theta_{cls}^j v_i}}{\sum_{k \in C} e^{\theta_{cls}^k v_i}}
\end{equation}

This approach ensures the unified classifier is trained on representative feature samples, effectively mitigating catastrophic forgetting while maintaining performance across tasks.

\subsection{Overall Optimization}
As shown in Fig. \ref{fig3}, our analysis identifies three distinct aspects of forgetting, each addressed through tailored loss functions. Throughout continual learning, the pre-trained temporal model parameters remain frozen, with only the Adapter, classifier, and DCN being updated. In the first stage of each task, a local cosine classifier is initialized and trained using \(\mathcal{L}_{cos}\) (Eq. \ref{deqn_e4}), while the old classification heads remain frozen. Knowledge distillation loss \(\mathcal{L}_{kd}\) (Eq. \ref{deqn_e3}) and drift compensation loss \(\mathcal{L}_{dc}\) (Eq. \ref{deqn_e5}) are also applied to constrain parameter updates and preliminarily train the DCN using feature drift between old and new models on new task samples.
In the second stage, the DCN is trained using \(\mathcal{L}_{dc}\) (Eq. \ref{deqn_e5}), with both the old model and the model trained in the first stage frozen. This stage focuses exclusively on DCN training, excluding the classification heads and other losses.
In the third stage, the DCN calibrates old class prototypes by projecting them into the new model's feature space, merging them with new class prototypes \(p^t\) (Eq. \ref{deqn_e6}) to form \(P_c\). Using \(P_c\) and their covariance matrices \(\Sigma_c\), a Gaussian distribution is formed, from which feature samples \(\mathcal{V}^t\) for old classes are sampled. The classification head is then retrained using \(\mathcal{L}_{ce}\) (Eq. \ref{deqn_e8}).
The loss functions for the three stages are:
\begin{equation}
\label{deqn_e9}
\begin{split}
\mathcal{L}_{s1} = \mathcal{L}_{cos} + \alpha \mathcal{L}_{kd} + \beta \mathcal{L}_{dc}, \\ \quad \mathcal{L}_{s2} = \mathcal{L}_{dc}, \quad \mathcal{L}_{s3} = \mathcal{L}_{ce} \quad
\end{split}
\end{equation}

where \(\alpha\) and \(\beta\) are weighting parameters balancing the loss terms. This three-stage approach effectively mitigates forgetting while maintaining performance across tasks.

\section{Experiments}
\subsection{Dataset}
We follow the experimental setup of [46], ensuring an approximately equal number of training samples per class. Based on this, we selected two time-series (TS) related applications: Human Activity Recognition (HAR) and Gesture Recognition. Typically, subjects/volunteers perform various activities or gestures within a fixed time frame. These datasets are suitable for Class-Incremental Learning (CIL) as each class exhibits a balanced sample distribution when tasks are divided. Although some studies have utilized HAR datasets for CIL\cite{re54,re55,re56}, they often use preprocessed feature vectors as input. In contrast, our experiments directly use raw time-series data, focusing on temporal patterns without normalization during preprocessing. This aligns with CIL requirements, as the complete dataset is inaccessible before training. Under this setup, we selected five commonly used open-source real-world time-series datasets for evaluation. Tab. \ref{tab2} provides an overview of these datasets. Includes the shape of each time-series data, the size of the training and test sets, the number of categories contained in that dataset, and the number of tasks in the experimental stream.

\subsubsection{UWave\cite{re57uwave}}
This dataset includes over 4,000 samples from 8 subjects, generating 8 simple gesture patterns. Data is collected from a 3-axis accelerometer, with each input sample comprising 315 time steps of 3-dimensional time-series data.

\subsubsection{UCI-HAR\cite{re58}}
Contains time-series data from smartphone inertial sensors during 6 different daily activities. Data is collected at 50Hz from 30 volunteers of varying ages. The raw time-series data, consisting of 9 channels over 128 time steps, is used as input.
\subsubsection{PAMAP2\cite{re59}}
A comprehensive physical activity monitoring dataset recording 12 different activities, such as walking, cycling, and playing soccer. Data is collected from 9 subjects wearing 3 IMU sensors and 1 heart rate monitor. We downsample the data from 100Hz to 33.3Hz and extract samples using a sliding window of size 170 with 50\% overlap. The dataset is split into training and testing sets at a 3:1 ratio, ensuring both sets include data from all subjects.

\subsubsection{DSA\cite{re60}}
Captures motion sensor data from 8 volunteers performing 19 daily sports activities. Each segment is treated as a sample, recorded over 45 channels with 125 time steps. To ensure class balance, we select 18 categories from this dataset for experiments.

\subsubsection{WISDM\cite{re61}}
A sensor-based human activity recognition dataset containing 18 activities involving 51 subjects. Following\cite{re62}, we use smartphone accelerometer data and extract samples using a non-overlapping sliding window of 200 time steps. Each sample represents 10 seconds of time-series data at a sampling rate of 20Hz. The dataset is split into training and testing sets at a 3:1 ratio, ensuring both sets include data from all subjects.

These datasets provide a robust foundation for evaluating our approach in real-world CIL scenarios.

\begin{table}
\setlength{\belowcaptionskip}{0cm}
\setlength{\tabcolsep}{3.4pt} 
\begin{center}
\caption{Overview of The Benchmark Datasets.}
\label{tab2}
\begin{tabular}{@{}lccccc@{}}
\toprule
Dataset & \multicolumn{1}{l}{shape ($C$ × $L$)} & \multicolumn{1}{l}{Train Size} & \multicolumn{1}{l}{Test Size} & \multicolumn{1}{l}{Classes} & \multicolumn{1}{l}{Exp Tasks} \\ \midrule
UWave   & 3 × 315                             & 896                            & 3582                          & 8                           & 4                             \\
UCI-HAR & 9 × 128                             & 7352                           & 2947                          & 6                           & 3                             \\
PAMAP2  & 36 × 170                            & 6434                           & 1050                          & 12                          & 3                             \\
DSA     & 45 × 125                            & 6840                           & 2280                          & 18                          & 6                             \\
WISDM   & 3 × 200                             & 18184                          & 6062                          & 18                          & 6                             \\ 
\bottomrule
\end{tabular}
\end{center}
\vspace{-0.5cm}
\end{table}

We also observe that the Moment method, which processes channels independently, achieves strong performance on datasets with many channels but incurs longer training times. For datasets with a high number of channels (e.g., DSA with 45 channels and PAMAP2 with 36 channels), dimensionality reduction can significantly shorten training time with only a slight accuracy trade-off. Thus, we apply PCA to reduce the feature dimensions to one-third of the original (DSA to 15 channels and PAMAP2 to 12 channels) before inputting the data into the model. To comply with the CIL setting, where only current task samples are accessible, we perform PCA dimensionality reduction only when loading the current task's data, rather than during preprocessing. This ensures adherence to incremental learning constraints while optimizing computational efficiency.

\subsection{Comparison Methods}
Our method is compared with two traditional NECIL approaches: LWF\cite{lwf} and DT2W\cite{re43}. We also include five PTMs-based methods in our comparison: SLCA\cite{re17slca}, SimpleCIL\cite{re49}, Adam-ssf\cite{re49}, Adam-adapter\cite{re49}, and SSIAT\cite{re51}. Furthermore, we compare against two replay-based approaches, ASER\cite{re63} and CLOPS\cite{re37}. To ensure a fair comparison, all methods utilize the same pre-trained Moment\cite{re24moment} as their backbone. In addition, we conduct experiments with LWF and DT2W using a randomly initialized CNN backbone to highlight the benefits of pre-training compared to traditional from-scratch training strategies in CIL. Finally, we consider two special experimental settings: joint-train and fine-tuning. Joint-train refers to training a single model on all data at once, serving as an upper-bound performance estimate. Fine-tuning involves directly fine-tuning the model without employing any anti-forgetting algorithms. The implementation details for all compared methods are summarized in Tab. \ref{tab3}.

\subsection{Evaluation Metrics}
\label{sec:mereics}
We employ three standard metrics for Class-Incremental Learning (CIL) evaluation, along with an additional metrics to assess the effectiveness of our Drift Compensation Network (DCN). Let \(a_{i,j}\) denote the accuracy evaluated on the test set of task \(j \leq i\) after training on task \(i\). These metrics provide a comprehensive evaluation of model performance and stability across incremental tasks.
\subsubsection{Average Accuracy}
defined as \(\mathcal{A}_i = \frac{1}{i} \sum_{j=1}^{i} a_{i,j}\), represents the mean accuracy across all test sets up to task \(i\). This metric reflects the overall performance of the model, providing a comprehensive measure of its ability to retain knowledge and generalize across incremental tasks.
\subsubsection{Average Forgetting\cite{re64}}
\(\mathcal{F}_i = \frac{1}{i-1} \sum_{j=1}^{i-1} f_{i,j}\), where \(f_{i,j} = \max_{k \in \{1, \ldots, i-1\}} (a_{k,j}) - a_{i,j}\), quantifies the performance degradation of task \(j\) after learning task \(i\). This metric reflects the extent to which the model forgets previously learned knowledge when acquiring new tasks, providing insight into the stability of the model during incremental learning.
\subsubsection{Average Learning Accuracy\cite{re65}}
\(\mathcal{A}_{\textit{cur}} = \frac{1}{T} \sum_{i=1}^{T} a_{i,i}\), represents the average accuracy of the current task during the learning process. To reflect the final performance, the Average Accuracy \(\mathcal{A}_T\) and the Average Forgetting \(\mathcal{F}_T\) are typically reported. These metrics are computed after completing the final task and provide a comprehensive evaluation of the model's overall performance and stability across all tasks.

\subsubsection{Prototype distance}
Following the completion of task \(t\), the discrepancy between the updated prototypes and the ground truth is quantified. The prototypes retained from task \(t-1\) are updated using the DCN. The distance metric, \(D^t\), is formulated as
\begin{equation}
\label{deqn_e10}
D^t = \| P_{\text{updated}}^{t-1} - P_{\text{real}}^t \|_2 = \sum_{i=1}^{n} \left( P_{\text{updated}}^{t-1}(i) - P_{\text{real}}^t(i) \right)^2
\end{equation}
Here, \(i\) represents the index of the stored prototypes, \(P_{\text{updated}}^{t-1} = d^t(P^{t-1})\) denotes the prototypes updated by the DCN from task \(t-1\), and \(P_{\text{real}}^t\) are the true prototypes extracted by the feature extractor \(\mathcal{F}^t\) from \(\mathcal{D}^{t-1}\). This metric intuitively reflects the effectiveness of different prototype update strategies.


\subsection{Implementation Details}


To adhere to the standard CIL definition, we split the datasets into \(T\) tasks, ensuring each task contains mutually exclusive classes. Before splitting, we shuffle the class order and evenly distribute all classes across tasks, assigning two distinct classes per task. For datasets with a small number of tasks (e.g., UCI-HAR and UWave, with 3 and 4 tasks, respectively), we follow the first protocol\cite{re1414}, dividing each task into training, validation, and test sets. We then perform a grid search to select optimal hyperparameters based on validation performance across all tasks. For datasets with more tasks, we employ another protocol \cite{re64}, splitting tasks into a "validation" stream (for cross-validation and hyperparameter tuning) and an "experimental" stream (for training and evaluation). The validation stream contains 3 tasks. This approach aligns better with the CIL definition as it avoids requiring access to the entire task stream.

All experiments are conducted independently with three different random seeds. For each run, hyperparameters are tuned following the protocols described above, and the best configuration is selected. The final results are averaged over the three runs. We set the initial learning rate to 0.005 and the batch size to 16. All model parameters are optimized using the SGD optimizer with the OneCycleLR scheduler for dynamic learning rate adjustment. All experiments are implemented using PyTorch on a single NVIDIA 4090 GPU.

\begin{table}
\setlength{\abovecaptionskip}{-0.1cm} 
\setlength{\belowcaptionskip}{0cm}
\setlength{\tabcolsep}{2.9pt} 
\begin{center}
\caption{Summary of The Comparison CIL Algorithms.}
\label{tab3}
\begin{tabular}{@{}l|ccc@{}}
\bottomrule
\multicolumn{1}{c|}{Algorithm} & Application & Category       & Characteristics     \\ \hline
LwF\cite{lwf}                            & Visual      & Regularization & KD on logits        \\
DT2W\cite{re43}                           & Time Series & Regularization & KD on feature maps  \\ \hline
ASER\cite{re63}                           & Visual      & Experience     & Memory retrieval    \\
CLOPS\cite{re37}                          & Time Series & Experience     & Update \& Retrieval \\ \hline
SLCA\cite{re17slca}                      & Visual      & PTMs           &Learning rates                     \\
SimpleCIL\cite{re49}                      & Visual      & PTMs           &Classifiers \& Prototype                      \\
Adam-ssf\cite{re49}                       & Visual      & PTMs           &Adaptation \& Branch fusion                     \\
Adam-adapter\cite{re49}                   & Visual      & PTMs           &Adaptation \& Branch fusion                      \\
SSIAT\cite{re51}                          & Visual      & PTMs           &Adaptation \& Semantic shift                     \\ \toprule 
\end{tabular}
\end{center}
\vspace{-0.5cm}
\end{table}


\subsection{Experiments Results}
\begin{table*}
\setlength{\abovecaptionskip}{0cm} 
\setlength{\belowcaptionskip}{0cm}
\setlength{\tabcolsep}{4.1pt} 
\begin{center}
\caption{Comparison with Different Incremental Learning Methods on Various Dataset Settings. Metrics Introduced in Section \ref{sec:mereics} Are Reported, Which Are \textbf{\(\mathcal{A}_T\)(↑)}, \textbf{\(\mathcal{F}_T\)(↓)} And \textbf{\(\mathcal{A}_{\textit{cur}}\)(↑)}. For Each Metric, Its Mean And Confidence Interval on 3 runs are reported}
\label{tab4}
\begin{tabular}{c|c|cc|cccc|cccccc}


\bottomrule
\multirow{2}{*}{Dataset} & \multirow{2}{*}{Metric} & Joint        &fine-tuning     & \multicolumn{2}{c}{LwF}   & \multicolumn{2}{c|}{DT2W} &SLCA        &SimpleCIL   &Adam-ssf         &Adam-adapter     &SSIAT       &ours       \\ \cline{3-14} 
                         &                         & \multicolumn{2}{c|}{Moment} & CNN         & Moment      & CNN         & Moment      & \multicolumn{6}{c}{Moment}                                                        \\ \hline
\multirow{3}{*}{UWave}   &\(\mathcal{A}_T\)                         & 93.9\textnormal{\tiny ±0.4}     & 24.5\textnormal{\tiny ±0.9}   & 32.2\textnormal{\tiny ±7.6}  & 44.5\textnormal{\tiny ±6.9}  & 56.1\textnormal{\tiny ±8.1}  & 65.2\textnormal{\tiny ±20.4} & 72.7\textnormal{\tiny ±22.4} & 83.1\textnormal{\tiny ±0.5}  & 83\textnormal{\tiny ±0.5}  & 83.2\textnormal{\tiny ±0.2}  & 73.9\textnormal{\tiny ±6.3}  & \textbf{85.1\textnormal{\tiny ±9.7}}  \\ 
                         &\(\mathcal{F}_T\)                         & N.A& 98.1\textnormal{\tiny ±0.6}& 61.2\textnormal{\tiny ±37.2}& 21.2\textnormal{\tiny ±12.3}& 33.3\textnormal{\tiny ±23.1}& 21.4\textnormal{\tiny ±7.8}& 19.4\textnormal{\tiny ±7.2}& 5.2\textnormal{\tiny ±7}& 5.4\textnormal{\tiny ±7.1}& 5.2\textnormal{\tiny ±7}& \textbf{4.5\textnormal{\tiny ±2.6}}&             11.1\textnormal{\tiny ±15.7}\\ 
                         &\(\mathcal{A}_{\textit{cur}}\)                         & N.A& 98.2\textnormal{\tiny ±1.1}& 78.1\textnormal{\tiny ±21.3}& 60.3\textnormal{\tiny ±26.2}& 77.4\textnormal{\tiny ±20.4}& 80.3\textnormal{\tiny ±23.4}& 86.3\textnormal{\tiny ±7.9}& 87.0\textnormal{\tiny ±5.4}& 86.9\textnormal{\tiny ±4.9}& 87.1\textnormal{\tiny ±5.4}& 65.9\textnormal{\tiny ±3.8}&             \textbf{93.4\textnormal{\tiny ±5.6}} \\ \hline
\multirow{3}{*}{UCI-HAR} &\(\mathcal{A}_T\)                          & 94.7\textnormal{\tiny ±0.2}   & 32.9\textnormal{\tiny ±6.9}   & 45.8\textnormal{\tiny ±14.1} & 67.4\textnormal{\tiny ±17.2} & 80.2\textnormal{\tiny ±6.0}  & 79.1\textnormal{\tiny ±13.1}  & 64.4\textnormal{\tiny ±16.9}  & 85.7\textnormal{\tiny ±0.6}   & 85.9\textnormal{\tiny ±0.6}  & 83.6\textnormal{\tiny ±7.6}   & 51.1\textnormal{\tiny ±7.3}& \textbf{88.4\textnormal{\tiny ±10.3}} \\ 
                         &\(\mathcal{F}_T\)                          & N.A& 97.7\textnormal{\tiny ±6.2}& 77.7\textnormal{\tiny ±17.8}& 16.6\textnormal{\tiny ±12.3}& 9.2\textnormal{\tiny ±4.3}& 16.1\textnormal{\tiny ±10.0}& 8.4\textnormal{\tiny ±6.6}& 5.7\textnormal{\tiny ±1}& \textbf{5.6\textnormal{\tiny ±0.7}}& 7.2\textnormal{\tiny ±3.6}& 62.8\textnormal{\tiny ±12.1}&             10.5\textnormal{\tiny ±25.3}\\ 
                         &\(\mathcal{A}_{\textit{cur}}\)                         & N.A& 98.3\textnormal{\tiny ±3.8}& 97.6\textnormal{\tiny ±3.3}& 78.4\textnormal{\tiny ±1267}& 90.9\textnormal{\tiny ±1.0}& 84.1\textnormal{\tiny ±14.0}& 66.0\textnormal{\tiny ±18.2}& 89.5\textnormal{\tiny ±0.5}& 89.5\textnormal{\tiny ±1.2}& 88.3\textnormal{\tiny ±5.4}& 92.9\textnormal{\tiny ±2.1}&             \textbf{95.0\textnormal{\tiny ±6.2}} \\ \hline
\multirow{3}{*}{pamap2}  &\(\mathcal{A}_T\)                          & 94.1\textnormal{\tiny ±2.5}    & 33.1\textnormal{\tiny ±0.3}   & 38.8\textnormal{\tiny ±17.0} & 73.5\textnormal{\tiny ±11.1} & 57.5\textnormal{\tiny ±12.7} & 85.7\textnormal{\tiny ±18.3} & 69.4\textnormal{\tiny ±21.0} & 91.1\textnormal{\tiny ±8.0}   & 90\textnormal{\tiny ±7.9}  & 91.3\textnormal{\tiny ±8.4}  & 82.8\textnormal{\tiny ±10.4}  & \textbf{92.7\textnormal{\tiny ±5.2}}  \\ 
                         &\(\mathcal{F}_T\)                          & N.A& 97.9\textnormal{\tiny ±1.6}& 71.4\textnormal{\tiny ±24.7}& 21.6\textnormal{\tiny ±29.9}& 35.2\textnormal{\tiny ±14.1}& 14.5\textnormal{\tiny ±14.4}& 30.6\textnormal{\tiny ±10.4}& 3.3\textnormal{\tiny ±2.7}& 3.6\textnormal{\tiny ±4.0}& 3.6\textnormal{\tiny ±2.7}& 20.1\textnormal{\tiny ±10.1}&             \textbf{3.2\textnormal{\tiny±4.1}} \\ 
                         &\(\mathcal{A}_{\textit{cur}}\)                         & N.A& 98.3\textnormal{\tiny ±1.0}& 86.4\textnormal{\tiny ±5.8}& 87.9\textnormal{\tiny ±9.1}& 80.9\textnormal{\tiny ±13.9}& 85.8\textnormal{\tiny ±11.3}& 91.1\textnormal{\tiny ±7.4}& 92.4\textnormal{\tiny ±4.4}& 92.4\textnormal{\tiny ±4.4}& 93.7\textnormal{\tiny ±5.8}& 96.2\textnormal{\tiny ±2.9}&             \textbf{95.6\textnormal{\tiny ±2.5}} \\ \hline
\multirow{3}{*}{DSA}     &\(\mathcal{A}_T\)                          & 98.8\textnormal{\tiny ±0.2}    & 16.7\textnormal{\tiny ±0.2}   & 19.3\textnormal{\tiny ±3.0}  & 51.5\textnormal{\tiny ±33.0} & 20.3\textnormal{\tiny ±8.3}  & 82.3\textnormal{\tiny ±17.8} & 62.8\textnormal{\tiny ±9.3}  & 90.7\textnormal{\tiny ±7.1}  & 90.7\textnormal{\tiny ±6.7}  & 90.4\textnormal{\tiny ±7.4}  & 47.9\textnormal{\tiny ±10.2} & \textbf{96.8\textnormal{\tiny ±6.5}}  \\ 
                         &\(\mathcal{F}_T\)                          & N.A& 98.9\textnormal{\tiny ±1.1}& 92.6\textnormal{\tiny ±10.2}& 35.1\textnormal{\tiny ±21.7}& 94.3\textnormal{\tiny ±12.5}& 20.0\textnormal{\tiny ±18.6}& 43.9\textnormal{\tiny ±10.0}& 4.4\textnormal{\tiny ±4.2}& 4.3\textnormal{\tiny ±4.2}& 4.7\textnormal{\tiny ±4.5}& 61.9\textnormal{\tiny ±12.4}&             \textbf{2.8\textnormal{\tiny±1.7}} \\ 
                         & \(\mathcal{A}_{\textit{cur}}\)                & N.A& 99.6\textnormal{\tiny ±0.2}& 96.4\textnormal{\tiny ±8.1}& 80.7\textnormal{\tiny ±16.9}& 98.8\textnormal{\tiny ±2.2}& 98.9\textnormal{\tiny ±1.2}& 93.5\textnormal{\tiny ±4.1}& 93.3\textnormal{\tiny ±6.2}& 94.2\textnormal{\tiny ±4.2}& 94.4\textnormal{\tiny ±4.0}& 98.3\textnormal{\tiny ±4.2}&             \textbf{99.5\textnormal{\tiny ±0.4}} \\ \hline
\multirow{3}{*}{WIDSM}   &\(\mathcal{A}_T\)                          & 53.3\textnormal{\tiny ±12.4}  & 14.0\textnormal{\tiny ±7.3}   & 18.5\textnormal{\tiny ±3.4}  & 19.4\textnormal{\tiny ±7.9}  & 15.3\textnormal{\tiny ±7.3}  & 25.6\textnormal{\tiny ±12.1} & 24.0\textnormal{\tiny ±10.2} & 33.5\textnormal{\tiny ±12.3} & 33.3\textnormal{\tiny ±12.3} & 31.8\textnormal{\tiny ±10.6} & 22.6\textnormal{\tiny ±13.3} & \textbf{36.1\textnormal{\tiny ±8.6}}   \\ 
                         &\(\mathcal{F}_T\)                          & N.A& 86.6\textnormal{\tiny ±13.9}& 86.3\textnormal{\tiny ±3.4}& 23.1\textnormal{\tiny ±11.1}& 35.8\textnormal{\tiny ±19.1}& 16.7\textnormal{\tiny ±7.6}& 67.1\textnormal{\tiny ±21.9}& 20.2\textnormal{\tiny ±9.5}& 20.4\textnormal{\tiny ±9.6}& \textbf{20\textnormal{\tiny ±9.2}}& 69.2\textnormal{\tiny ±14.5}&             30.7\textnormal{\tiny ±5.9}\\ 
                         &\(\mathcal{A}_{\textit{cur}}\)                  & N.A& 87.2\textnormal{\tiny ±1.3}& 90.4\textnormal{\tiny ±5.4}& 38.4\textnormal{\tiny ±22.6}& 43.4\textnormal{\tiny ±18.0}& 31.8\textnormal{\tiny ±8.3}& 79.9\textnormal{\tiny ±19.9}& 50.4\textnormal{\tiny ±19.9}& 50.4\textnormal{\tiny ±20.1}& 48.4\textnormal{\tiny ±19}& \textbf{80.3\textnormal{\tiny ±11.1}}&             61.5\textnormal{\tiny ±11.8}\\  \toprule
\end{tabular}
\end{center}
\vspace{-0.5cm}
\end{table*}

\begin{figure*}[!t]
    \centering
    \begin{minipage}{\textwidth}
        \centering
        \includegraphics[width=1.0\textwidth]{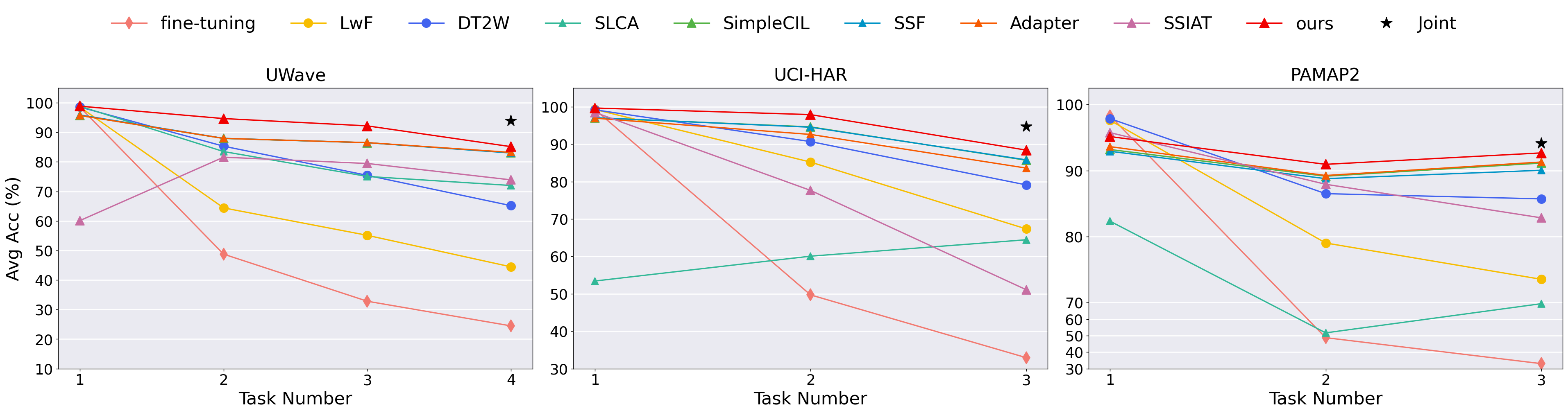} 
    \end{minipage}\\[3pt] 
    \begin{minipage}{\textwidth}
        \centering
        \includegraphics[width=0.75\textwidth]{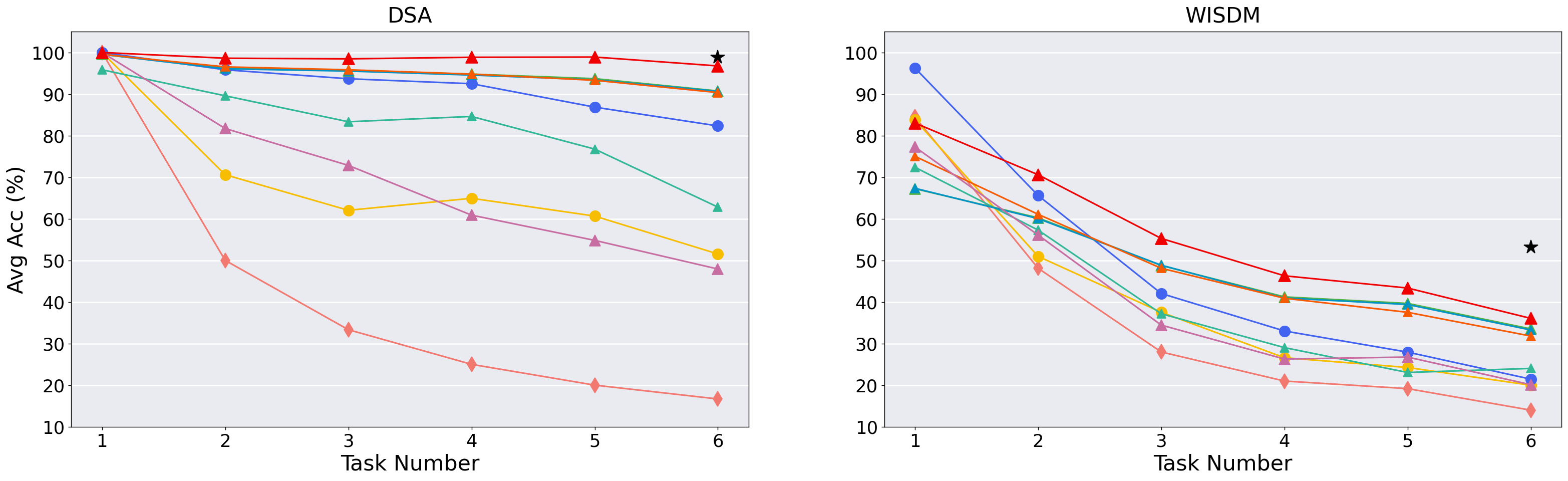} 
    \end{minipage}
    \caption{Evolution of Average Accuracy (\(\mathcal{A}_i\)) is shown. Traditional CIL methods use circular markers, while PTM-based methods use triangular markers. Since Joint represents joint-train across the entire task sequence, its result is shown as a single point rather than a curve.} 
    \label{fig4} 
\vspace{-0.5cm}
\end{figure*}

Tab. \ref{tab4} shows the main results. 
When \textbf{compared to traditional methods}, an analysis under two architectures (CNN and pretrained Moment) reveals two critical findings: (1) Pretrained Moment significantly enhances traditional CIL performance, achieving a 21.3\%-37.8\% accuracy gain over CNN baselines, validating its foundational value; (2) DT2W\cite{re43}, a regularization method specifically designed for time series data, outperforms the traditional regularization approach LWF\cite{lwf} by 8.7\%-15.2\% with Moment. However, our method still surpasses DT2W by 7.0\%-19.9\% across five datasets.


\textbf{Compared to PTM-based methods}, our method achieves higher final accuracy than other PTM-based approaches (SLCA\cite{re17slca}, SimpleCIL\cite{re49}, Adam ssf\cite{re49}, Adam-adapter\cite{re49}, SSIAT\cite{re51}) by balancing plasticity and stability. On the UWave dataset, while SimpleCIL, Adam-ssf, and Adam-adapter exhibit superior anti-forgetting performance (forgetting rates: 5.2\%, 5.4\%, 5.2\% vs. our 11.1\%), their limited plasticity results in lower learning accuracy (87.0\%, 86.9\%, 87.1\% vs. our 93.4\%), leading to 1.9\%-2.1\% lower final accuracy than our method. This performance gap stems from their reliance on frozen PTM embeddings for classifier construction, which limits adaptation to diverse datasets. Furthermore, our method reduces forgetting rates by 16.9\%-59.1\% compared to SSIAT. This improvement is attributable to our drift compensation network’s ability to correct old-class prototypes, representing a significant advance over semantic drift mitigation strategies.

\textbf{Compared to Joint-train and fine-tuning}, our method reduces the performance gap with joint-train (the theoretical upper bound) to just 1.4\% and 2.0\% on the PAMAP2 and DSA datasets, demonstrating near-optimal continual knowledge integration. In contrast, fine-tuning without anti-forgetting mechanisms suffers severe catastrophic forgetting (86.6\%-98.9\%), highlighting the necessity of deliberate CIL design.


\textbf{Accuracy Curve}, 
As shown in Fig. \ref{fig4}, the evolution of average accuracy (\(\mathcal{A}_{i}\)) across the five datasets is comprehensively visualized.
Notably, while initial accuracy levels are similar, our method achieves the best results in subsequent stages. This observation highlights that our approach achieves a superior balance between mitigating forgetting and acquiring new knowledge.

\textbf{Compared to ER-based methods}, we further evaluated our method against two replay-based approaches, ASER\cite{re63} and CLOPS\cite{re37}, each retaining 5\% of samples per class and using Moment as the backbone. The experimental results, shown in Tab. \ref{tab5}, indicate that our method still holds a competitive edge. Specifically, it achieves performance improvements of 1.6\%, 2.8\%, and 3.7\% on the UWave, PAMAP2, and DSA datasets, respectively. These results underscore the effectiveness of our approach in mitigating forgetting, even when compared to replay-based methods.

\begin{table}
\setlength{\belowcaptionskip}{0cm}

\begin{center}
\caption{Comparison With Different Rplay-Based Methods on Various Dataset Settings. \ding{52} Represent Exemplar-free setting}
\label{tab5}
\begin{tabular}{@{}c|c|cc|c@{}}


\bottomrule
\multirow{2}{*}{Dataset} & \multirow{2}{*}{Metric} & ASER                & CLOPS               & ours     \\ \cline{3-5} 
                      &                         & \ding{55} & \ding{55} & \ding{52} \\ \hline

\multirow{3}{*}{UWave}   & \(\mathcal{A}_T\)      & 83.5\textnormal{\tiny ±6.2}  & 78.1\textnormal{\tiny ±4.9}  & \textbf{85.1\textnormal{\tiny ±9.7}}  \\
                         & \(\mathcal{F}_T\)      & 18.2\textnormal{\tiny ±9.3}  & 25.8\textnormal{\tiny ±6.9}  & \textbf{11.1\textnormal{\tiny ±15.7}} \\
                         & \(\mathcal{A}_{\textit{cur}}\)     & 97.2\textnormal{\tiny ±0.8}  & \textbf{97.4\textnormal{\tiny ±0.8}}  & 93.4\textnormal{\tiny ±5.6}  \\ \hline
\multirow{3}{*}{UCI-HAR} & \(\mathcal{A}_T\)      & \textbf{91.6\textnormal{\tiny ±4.4}}  & 85.7\textnormal{\tiny ±15.4} & 88.4\textnormal{\tiny ±10.3} \\
                         & \(\mathcal{F}_T\)      & \textbf{9.6\textnormal{\tiny ±4} }    & 18.1\textnormal{\tiny ±16.2} & 10.5\textnormal{\tiny ±25.3} \\
                         & \(\mathcal{A}_{\textit{cur}}\)     & 97.6\textnormal{\tiny ±2.9}  & \textbf{97.8\textnormal{\tiny ±4.4}}  & 95\textnormal{\tiny ±6.2}    \\ \hline
\multirow{3}{*}{pamap2}  & \(\mathcal{A}_T\)      & 88.5\textnormal{\tiny ±4.1}  & 89.9\textnormal{\tiny ±3.4}  & \textbf{92.7\textnormal{\tiny ±5.2} } \\
                         & \(\mathcal{F}_T\)      & 14.5\textnormal{\tiny ±6.7}  & 11.2\textnormal{\tiny ±2.5}  & \textbf{3.2\textnormal{\tiny ±4.1}}   \\
                         & \(\mathcal{A}_{\textit{cur}}\)     & \textbf{98.1\textnormal{\tiny ±0.9}}  & 97.4\textnormal{\tiny ±1.9}  & 95.6\textnormal{\tiny ±2.5}  \\ \hline
\multirow{3}{*}{DSA}     & \(\mathcal{A}_T\)      & 87.6\textnormal{\tiny ±13.8} & 93.1\textnormal{\tiny ±13.4} & \textbf{96.8\textnormal{\tiny ±6.5}}  \\
                         & \(\mathcal{F}_T\)      & 14.7\textnormal{\tiny ±16.1} & 8.3\textnormal{\tiny ±15.5}  & \textbf{2.8\textnormal{\tiny ±6.6}}   \\
                         & \(\mathcal{A}_{\textit{cur}}\)     & 99.1\textnormal{\tiny ±0.7}  & \textbf{99.9\textnormal{\tiny ±0.3}}  & 99.0\textnormal{\tiny ±1.7}  \\ \hline
\multirow{3}{*}{WIDSM}   & \(\mathcal{A}_T\)      & \textbf{36.7\textnormal{\tiny ±14.9}} & 35.8\textnormal{\tiny ±14.2} & 36.1\textnormal{\tiny ±8.6}  \\
                         & \(\mathcal{F}_T\)      & 54.5\textnormal{\tiny ±14.6} & 53.0\textnormal{\tiny ±14}   & \textbf{30.3\textnormal{\tiny ±6.0}}  \\
                         & \(\mathcal{A}_{\textit{cur}}\)     & \textbf{82.4\textnormal{\tiny ±12.8}} & 79.9\textnormal{\tiny ±18.3} & 61.5\textnormal{\tiny ±11.8} \\ \toprule
\end{tabular}
\end{center}
\vspace{-0.4cm}
\end{table}

\subsection{Ablation Study}
\subsubsection{Impact of Different Components on Final Performance}
To validate the effectiveness of our method, we conducted extensive experiments on the UWave, PAMAP2, and DSA datasets. Our approach consists of three components: a pre-trained model with adapter, Unified Classifier Training (UCT), and a Drift Compensation Network (DCN), with the Adapter serving as the base. The results, presented in Tab. \ref{tab6}, confirm the following observations:

(1) Unified Classifier Training (UCT) significantly enhances both model plasticity and resistance to forgetting, leading to substantial improvements in final accuracy. Specifically, accuracy improvements of 30.77\%, 25.27\%, and 48.69\% were observed on the three datasets, respectively. This can be attributed to UCT generating feature samples for each class and training the classifier at the end of each task, which not only consolidates knowledge of old classes but also improves the classifier's ability to distinguish between different classes, thereby enhancing plasticity.
(2) Drift Compensation Network (DCN) with UCT  further reduces forgetting and enhances plasticity, achieving additional improvements of 8.5\%, 4.1\%, and 3.3\% on the three datasets, respectively. This gain stems from the DCN learning the transformation function between the feature spaces of old and new models. Calibrated prototypes better represent historical data in the new model's feature space, enabling the sampled prototypes to generate more representative samples. This preserves more historical knowledge and mitigates forgetting.

\begin{table}
\setlength{\abovecaptionskip}{0cm} 
\setlength{\belowcaptionskip}{0cm}
\setlength{\tabcolsep}{3pt} 
\begin{center}
\caption{Ablation Study of Different Components.}
\label{tab6}
\begin{tabular}{l|lrl|lll|lll}
\bottomrule
\multicolumn{1}{c|}{\multirow{2}{*}{Method}} & \multicolumn{3}{c|}{UWave}                                                & \multicolumn{3}{c|}{PAMAP2}                                               & \multicolumn{3}{c}{DSA}                                                 \\  \cline{2-10}
\multicolumn{1}{c|}{}                        & \multicolumn{1}{c}{\(\mathcal{A}_T\)} & \multicolumn{1}{c}{\(\mathcal{F}_T\)} & \multicolumn{1}{c|}{\(\mathcal{A}_{\textit{cur}}\)} & \multicolumn{1}{c}{\(\mathcal{A}_T\)} & \multicolumn{1}{c}{\(\mathcal{F}_T\)} & \multicolumn{1}{c|}{\(\mathcal{A}_{\textit{cur}}\)} & \multicolumn{1}{c}{\(\mathcal{A}_T\)} & \multicolumn{1}{c}{\(\mathcal{F}_T\)} & \multicolumn{1}{c}{\(\mathcal{A}_{\textit{cur}}\)} \\ \hline
Base                                         & 45.8                   & 23.5                   & 61.9                    & 63.4                   & 12.4                   & 71.2                    & 44.8                   & 27.3                   & 67.6                   \\
Base+UCT                                     & 76.6                   & 19.0                   & 84.9                    & 88.7                   & 10.7                   & \textbf{95.6}                    & 93.5                   & 7.2                    & \textbf{99.5}                   \\ \rowcolor[HTML]{E7E7E8}
Base+UCT+DCN                                 & \textbf{85.1 }                  & \textbf{11.1 }                  & \textbf{93.4}                    & \textbf{92.8}                   & \textbf{3.2}                    & \textbf{95.6}                    & \textbf{96.8}                   & \textbf{2.8}                    & 99.0                   \\ \toprule
\end{tabular}
\vspace{-0.75cm}
\end{center}
\end{table}

\subsubsection{The impact of DCN training strategy}\label{sec:subDCN}
\begin{figure}[!t]
\begin{center}
\centering
\includegraphics[width=3.3in]{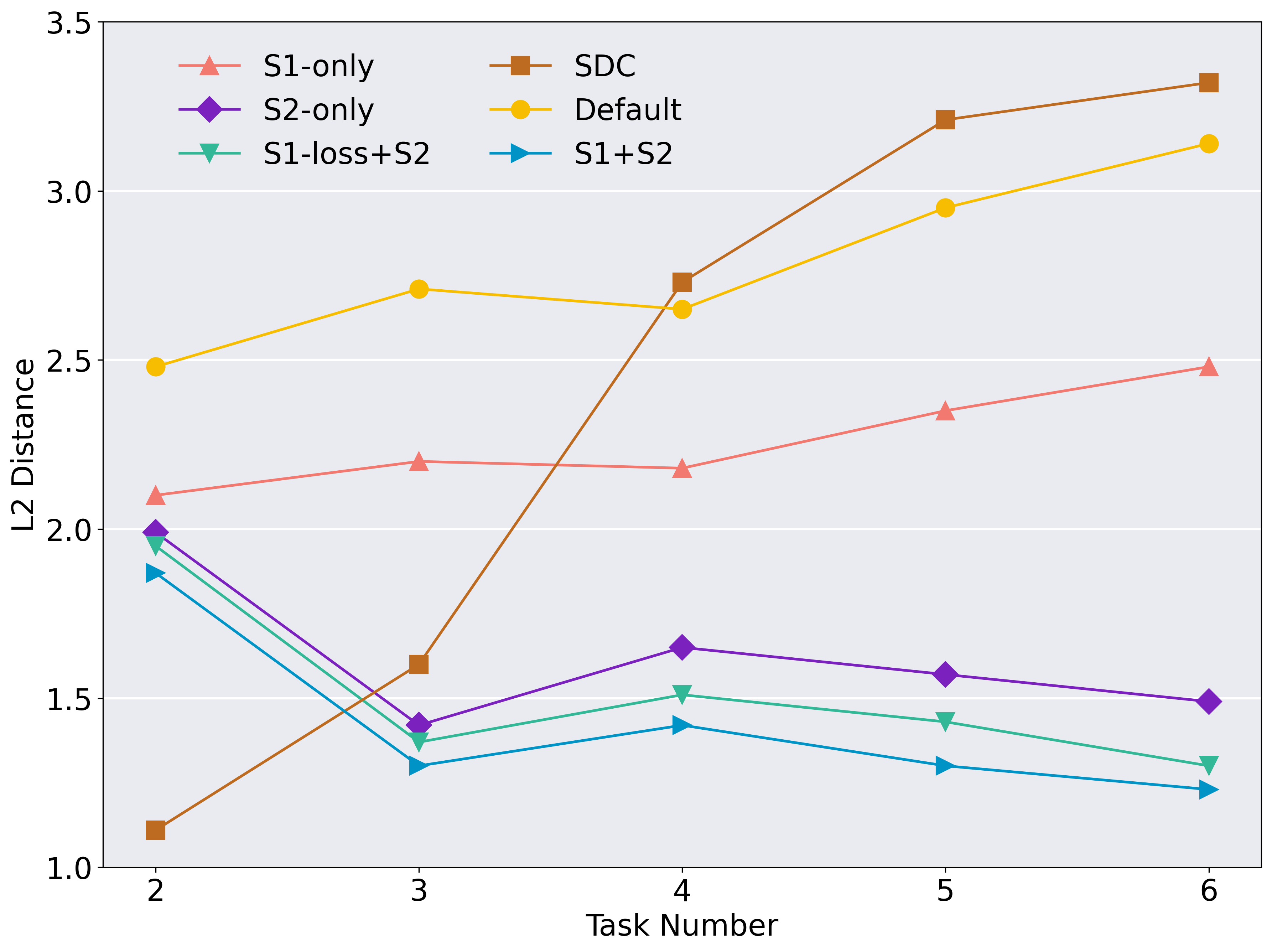}
\caption{Comparison of the L2 Distance Between Updated Prototypes and Real Prototypes Across Tasks for Different Strategies.}
\label{fig5}
\vspace{-0.4cm}
\end{center}
\end{figure}

To validate the effectiveness of the proposed two-stage DCN training strategy, we conducted experiments on the WISDM dataset. Starting from the second task, after updating the old classes prototypes using DCN, we calculated the L2 distance \( D^t \) (Eq. (\ref{deqn_e10})) between the updated old-class prototype \( P_{\text{updated}}^{t-1} \) and the ground-truth prototype \( P_{\text{real}}^t \) after each task. Comparative methods included \textbf{S1-only} (only perform the first stage of DCN training, as shown in Fig. \ref{fig3}(I)), \textbf{S2-only} (only perform the second stage of DCN training, as shown in Fig. \ref{fig3}(II)), \textbf{S1-loss+S2} (retains the \( \mathcal{L}_{dc} \) (Eq. (\ref{deqn_e5})) during first-stage adapter tuning,  while reinitializing and training the DCN from scratch in the second stage), \textbf{SDC} (semantic drift compensation, This strategy is used by SSIAT\cite{re51}), \textbf{Default} (Do not update the prototype), and \textbf{S1+S2} (the complete two-stage strategy we proposed), The experimental results shown in Fig. \ref{fig5}.

The results indicate that the \textbf{SDC} method exhibits the lowest \( D^t \) values in the initial tasks (Task 2-3), but its distance sharply increases starting from Task 4, exceeding the \textbf{Default} baseline by Task 5-6. indicating its prototype update mechanism causes reverse drift compensation over time. \textbf{Default} maintains high and rising \( D^t \), confirming static prototypes’ limitations. In contrast,  All DCN strategies suppress feature drift: \textbf{S1-only} limits drift growth modestly, while \textbf{S2-only} outperforms \textbf{S1-only} but lags behind \textbf{S1-loss+S2}, This highlights \( \mathcal{L}_{dc} \)’s (Eq. (\ref{deqn_e5})) role in guiding model parameter updates for efficient drift pattern learning. The full \textbf{S1+S2} strategy, combining joint model DCN optimization in the first stage and inherited DCN training in the second stage, achieves the lowest \( D^t \), demonstrating DCN's first-stage accumulated knowledge is progressively refined in the second stage, thereby strengthening its feature drift compensation capacity.

A deeper mechanism analysis showed that the introduction of \( \mathcal{L}_{dc} \) (Eq. (\ref{deqn_e5})) in the first stage constrained feature space shifts caused by model updates, laying the groundwork for learning drift patterns in the second stage DCN. The second stage then further enhanced the DCN's ability to model the relationship between features from new and old models based on the optimized feature distribution. The collaborative optimization mechanism of this two-stage drift compensation network effectively addresses the feature drift issue in old-class prototypes during class-incremental learning, providing an innovative solution to enhance model stability.

\subsubsection{The impact of different combinations of Adapter Scale Factor and KD weights}
\label{sec:kd}
\begin{figure}[!t]
\centering
\includegraphics[width=3.3in]{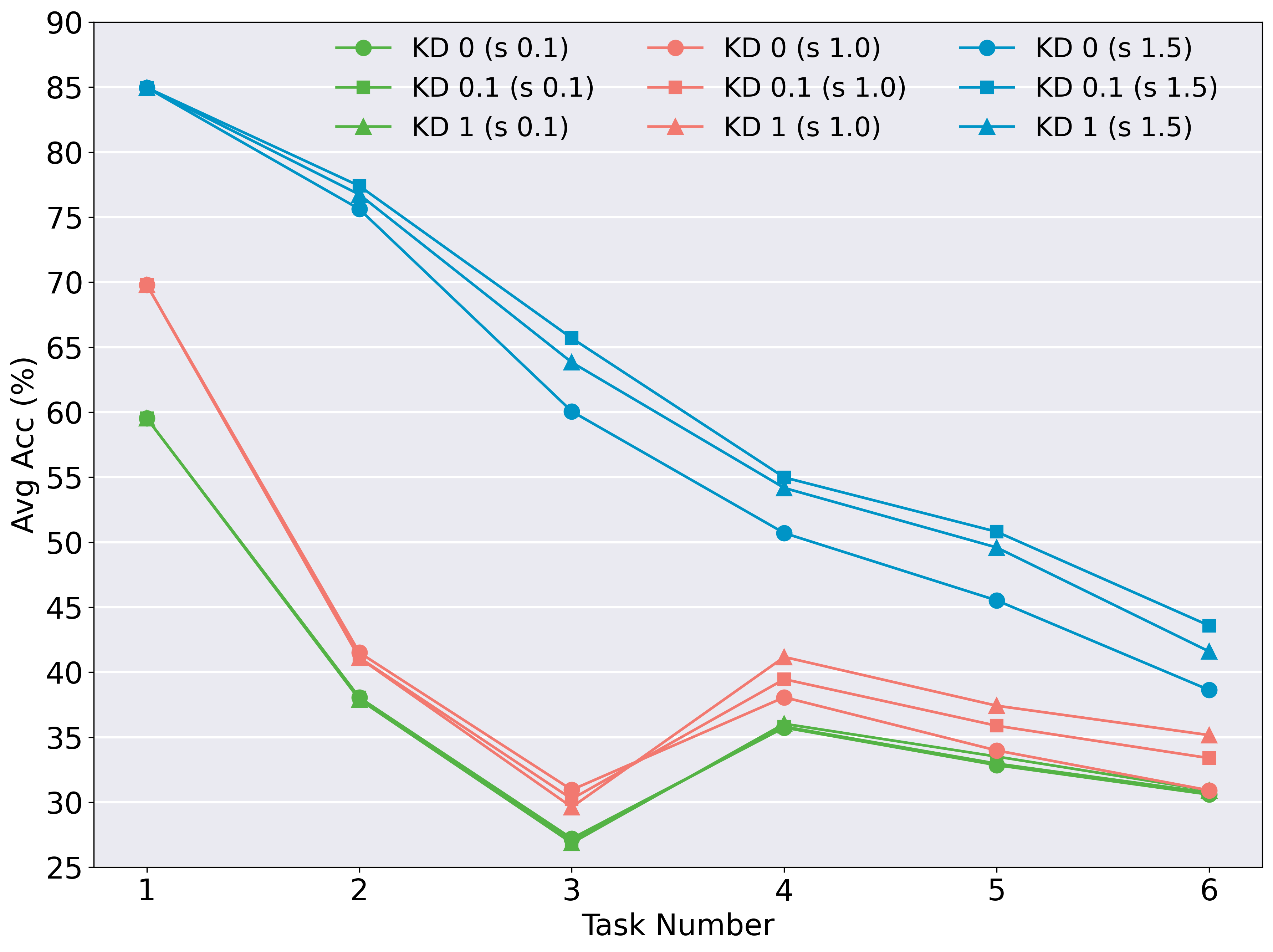}
\caption{Evolution of Average Accuracy (\(\mathcal{A}_i\)) of different combinations of Adapter Scale Factor and KD weights on the Widsm dataset.}
\label{fig6}
\end{figure}

In this study, we investigate the impact of different adapter scale factor (\(s\)) (Eq. (\ref{deqn_e2})) and Knowledge Distillation (KD) (Eq. (\ref{deqn_e3})) weight combinations. The experimental results, shown in Fig.  \ref{fig6}, reveal that when \(s\) is set to a small value, the accuracy of the initial task significantly decreases. Specifically, when \(s = 0.1\), the accuracy across tasks remains nearly unchanged regardless of the KD weight, as an excessively small \(s\) severely restricts model parameter updates, rendering KD ineffective. Conversely, assigning a larger \(s\) to the Adapter significantly enhances model plasticity, evidenced by higher accuracy in the initial stages of tasks. In this case, different KD weights have a more pronounced impact on accuracy. For instance, when \(s = 1.0\), a KD weight of 0.1 achieves the highest final accuracy, while when \(s = 1.5\), a KD weight of 1.0 yields the best performance.

The results demonstrate that, compared to SSIAT\cite{re51}, which uses a small adapter scale factor without KD, our approach better balances plasticity and stability. While SSIAT mitigates catastrophic forgetting to some extent, its limited plasticity is insufficient for time-series datasets, which often exhibit indistinct features and limited sample sizes. Broadly, \(s\) can be interpreted as a measure of model plasticity, while KD represents stability. By carefully balancing \(s\) and KD, our method achieves an optimal trade-off between plasticity and stability, leading to improved final accuracy.

\subsubsection{The Impact of Pre-trained Models Scales}
To evaluate how the scale of time series pre-trained models impacts the CIL performance of our method, we extend our original experiments (built upon the \textbf{MOMENT-base} backbone) by systematically comparing two additional variants (\textbf{MOMENT-small} and \textbf{MOMENT-large}) on the UWave, PAMAP2, and DSA datasets. As shown in Tab.\ref{tab7}, model scale exerts dataset-dependent effects on CIL: accuracy on UWave progressively increases with parameter count (from 82.27\% for small to 90.99\% for large), while \textbf{MOMENT-small} achieves optimal performance on both PAMAP2 (93.22\%) and \textbf{MOMENT-base} excels on DSA (96.76\%). Notably, \textbf{MOMENT-base} achieves holistically superior performance across three datasets, balancing accuracy and computational efficiency despite not being the largest variant.

\begin{table}
\setlength{\abovecaptionskip}{0cm} 
\setlength{\belowcaptionskip}{0cm}
\begin{center}
\caption{Final Accuracy (\(\mathcal{A}_T\)) Results for Different-Sized Pre-trained Moment Across Three Datasets.}
\label{tab7}
\begin{tabular}{ccccc}
\toprule
Size  & Parameters & UWave & PAMAP2 & DSA   \\ \midrule
small & 40M        & 82.27 & \textbf{93.22}  & 94.68 \\ \rowcolor[HTML]{E7E7E8}
base  & 125M       & 85.14 & 92.67  & \textbf{96.76} \\
large & 385M       & \textbf{90.99} & 91.92  & 96.28      \\ \bottomrule
\end{tabular}
\end{center}
\vspace{-0.4cm}
\end{table}

\section{Conclusion}
In recent years, pre-trained models (PTMs) have emerged as a powerful tool for class incremental learning (CIL), attracting significant attention in the research community. This paper revisits traditional CIL methodologies and pioneers the integration of PTMs into time series class incremental learning (TSCIL) scenarios. Our proposed framework combines shared adapter tuning with knowledge distillation (KD) and introduces a feature drift compensation network (DCN) to rectify the drift of old class prototypes. To enhance the effectiveness of DCN, we decompose its training process into two distinct steps, thereby strengthening its capability to mitigate prototype drift. Additionally, we leverage feature samples from the corrected prototypes generated by DCN to retrain the unified classifier at the final stage of each session. Without relying on exemplar retention, our method achieves an effective balance between model stability and plasticity. Extensive experiments on five benchmark datasets demonstrate the superiority of our approach, consistently achieving state-of-the-art performance. These findings provide novel insights into non-exemplar class-incremental learning, particularly in time series scenarios, and pave the way for future research in this direction.

\bibliography{reference.bib}{}
\bibliographystyle{IEEEtran}
\vspace{-1.4cm} 

\begin{IEEEbiography}[{\includegraphics[width=1in,height=1.25in,
		clip,keepaspectratio]{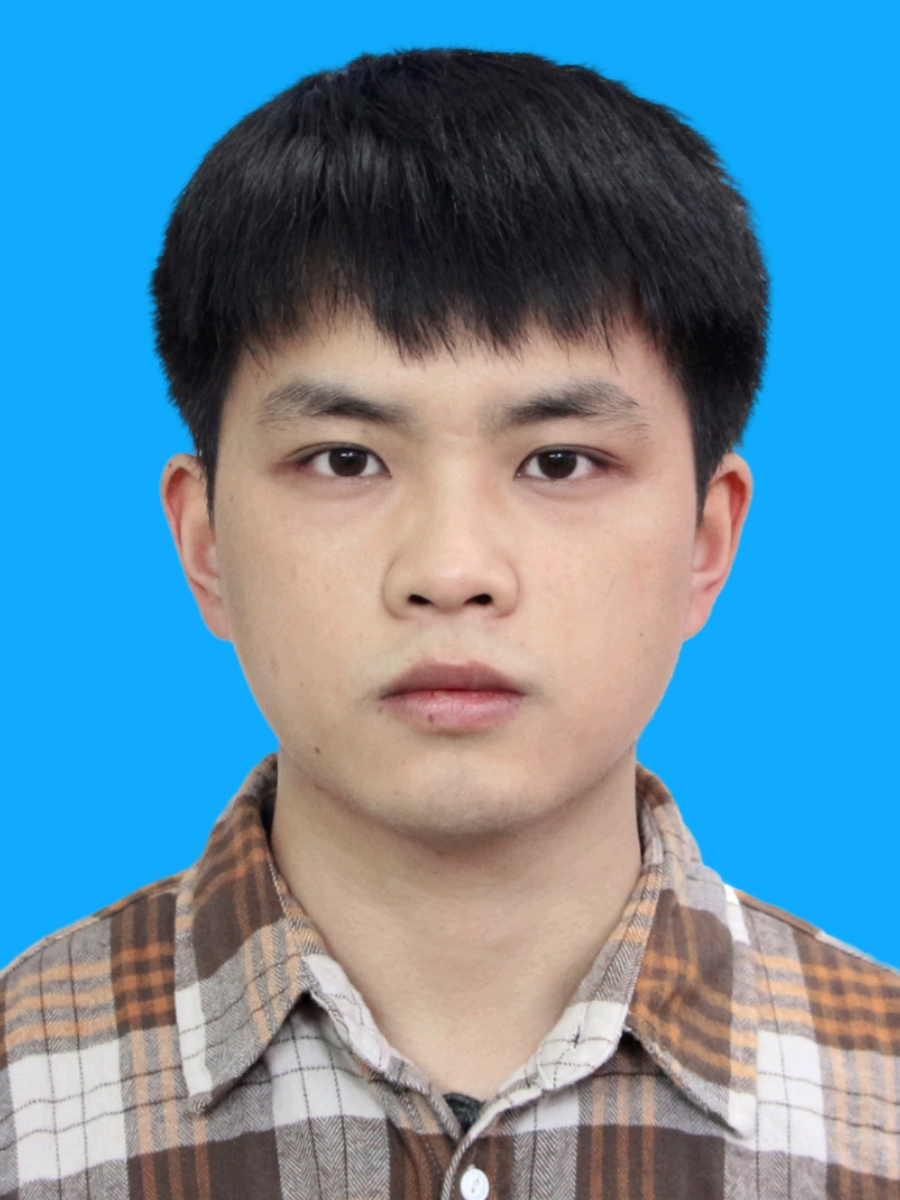}}]{Yuanlong Wu} 
	received B.S. degree from the School of Information and Communication Engineering, Dalian Minzu University, in 2023. He has been pursuing his Master’s degree in the School of Computer Science, University of South China since 2023. His current research interests include Human activity recognition and time series analysis.
\end{IEEEbiography}
\vspace{-1.4cm} 

\begin{IEEEbiography}[{\includegraphics[width=1in,height=1.25in,
		clip,keepaspectratio]{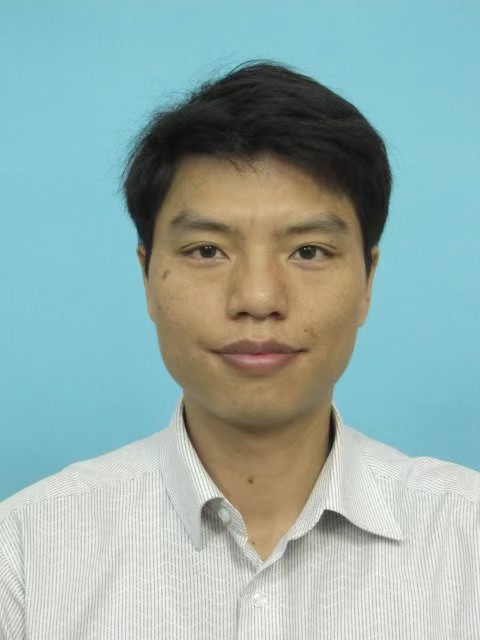}}]{MingxingNie} 
	obtained his Bachelor's degree in Computer Science from Central South University, Hunan, China, in 2004. Then he obtained his Master's degree in Computer system structure and PhD in Traffic information engineering and control both from Central South University, in 2007 and 2015. Currently, he is an associate professor at school of computer science, University of South China. His current research interests are Internet of Things and Optimization Technique.
\end{IEEEbiography}
\vspace{-1.4cm} 
\begin{IEEEbiography}[{\includegraphics[width=1in,height=1.25in,
		clip,keepaspectratio]{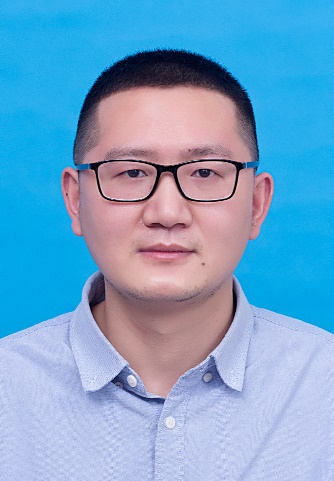}}]{Tao Zhu}
	received his B.E. degree from Central South University, Changsha, China, and Ph.D. from University of Science and Technology of China, Hefei, China, in 2009 and 2015 respectively. He is currently an associate professor at University of South China, Hengyang, China. He is the principal investigator of several projects funded by the National Natural Science Foundation of China and Science Foundation of Hunan Province etc. He is now the Chair of IEEE CIS Smart World Technical Committee Task Force on "User-Centred Smart Systems".His research interests include IoT, pervasive computing, assisted living and evolutionary computation.
\end{IEEEbiography}
\vspace{-1.4cm} 
\begin{IEEEbiography}[{\includegraphics[width=1in,height=1.25in,
		clip,keepaspectratio]{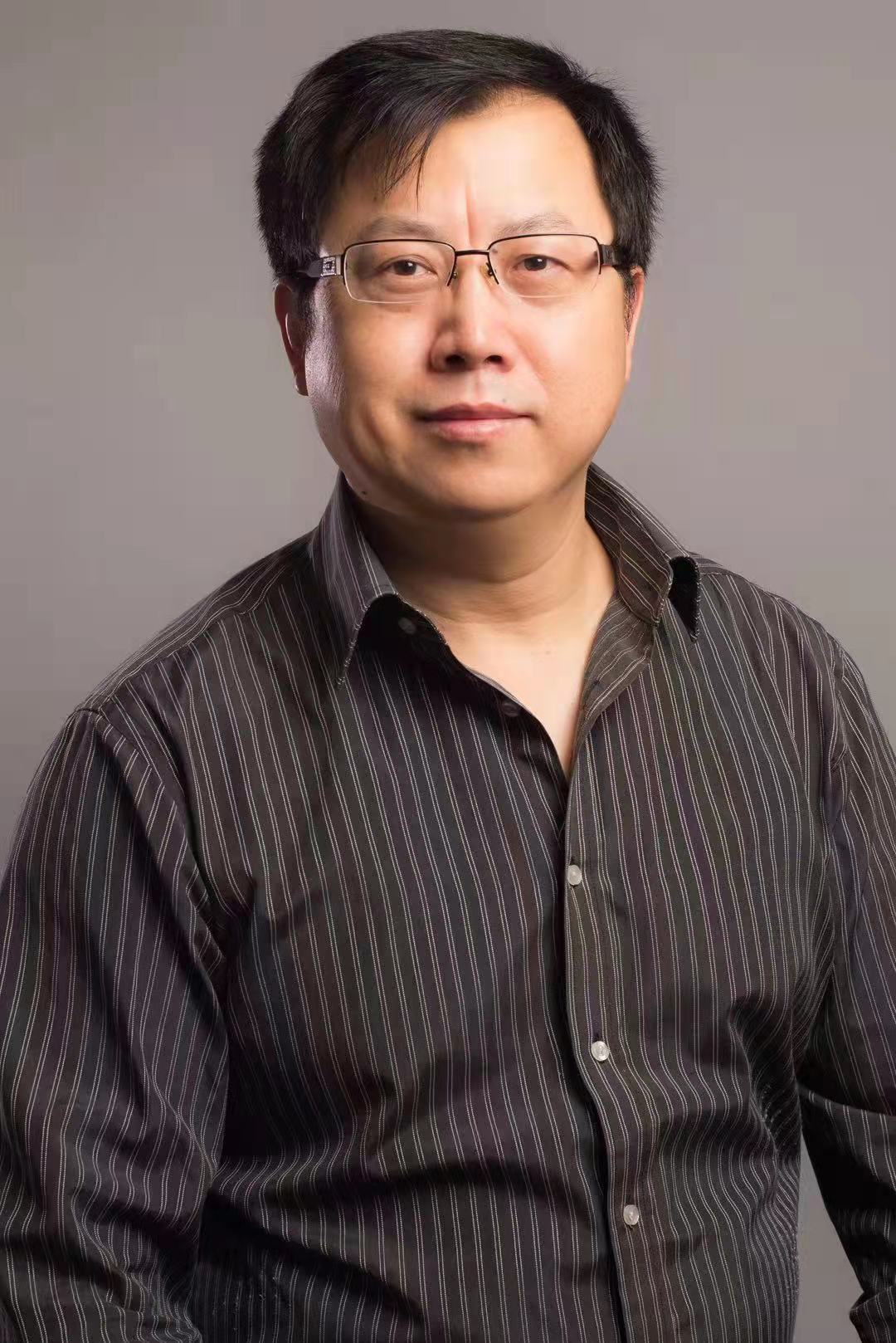}}]{Liming  Chen} 
	is a Professor of School of Computer Science and Technology, Dalian University of Technology, China. He holds a Doctoral Degree in computer Science from De Montfort University (1998.4 - 2001.11), a Master degree in Vehicle Engineering from Beijing Institute of Technology (1985.9 - 1988.3), and Bachelor Degree in Mechanical Engineering rom Beijing Institute of Technology (1981.9 - 1985.7), 
	His current research interests include pervasive computing, data analytics, artificial intelligence and user centered intelligent systems and their applications in health care and cybersecurity. He has published over 250 papers in the aforementioned areas.
	Liming is an IET Fellow and a Senior Member of IEEE.
\end{IEEEbiography}

\begin{IEEEbiography}[{\includegraphics[width=1in,height=1.25in,
		clip,keepaspectratio]{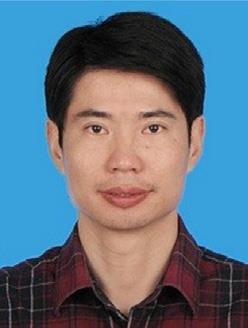}}]{Huansheng Ning}
	received his B.S. degree from
	Anhui University in 1996 and his Ph.D. degree
	from Beihang University in 2001. He is currently
	a Professor and Vice Dean with the School of
	Computer and Communication Engineering, University of Science and Technology Beijing and China
	and Beijing Engineering Research Center for Cyberspace Data Analysis and Applications, China,
	and the founder and principal at the Cybermatics and
	Cyberspace International Science and Technology
	Cooperation Base. He has authored several books
	and over 70 papers in journals and at international conferences/workshops.
	He has been the Associate Editor of the IEEE Systems Journal and IEEE 	Internet
	of Things Journal, Chairman (2012) and Executive Chairman (2013) of the
	program committee at the IEEE International Internet of Things Conference,
	and the Co-Executive Chairman of the 2013 International Cyber Technology
	Conference and the 2015 Smart World Congress. His awards include the
	IEEE Computer Society Meritorious Service Award and the IEEE Computer
	Society Golden Core Member Award. His current research interests include
	the 	Internet of Things, Cyber Physical Social Systems, electromagnetic sensing
	and computing.
\end{IEEEbiography}
\vspace{-7cm}

\begin{IEEEbiography}[{\includegraphics[width=1in,height=1.25in,
		clip,keepaspectratio]{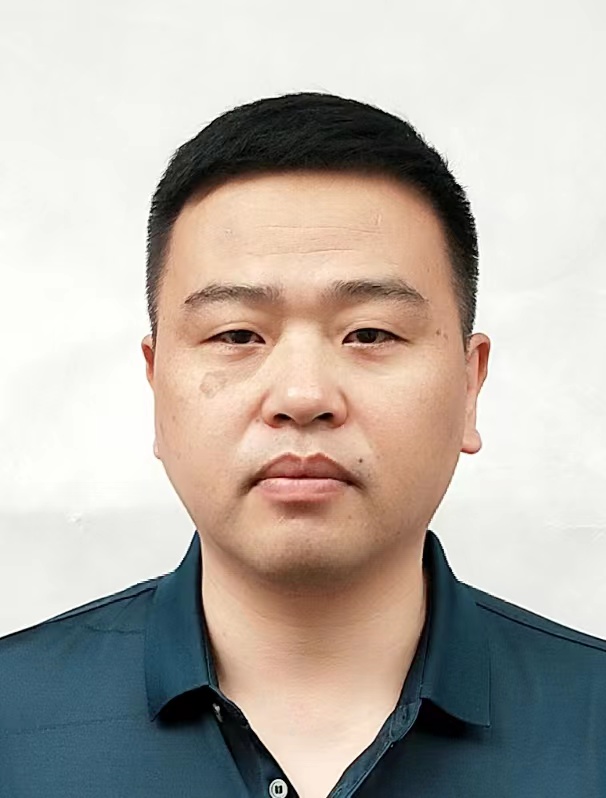}}]{Yaping Wan} received his B.S. degree from Huazhong University of Science\&Techno	logy(HUST) in 2004 and his Ph.D. degree from HUST in 2009. He is currently a Professor and Dean with the School of Computer, University of South China and the International Cooperation Research Center for Medical Big Data of Hunan Province. He has authored several books and over 40 papers in journals and at international conferences/workshops. He has been the Workshop Chairman (2022) at the 16th IEEE International Conference on Big Data Science and Engineering, and the Session Chairman (2021,2022) of Asian Conference on Artificial Intelligence Technology.  His current research interests include intelligent nuclear security, big data analysis and causal inference, high-reliability computing and security evaluation.
	\vspace{28 mm} 
\end{IEEEbiography}

\end{document}